\documentclass[11pt,letterpaper]{article}

\usepackage[utf8]{inputenc}
\usepackage{amsmath}
\usepackage{amsfonts}
\usepackage{amssymb}
\usepackage[left=2cm,right=2cm,top=2cm,bottom=2cm]{geometry}
\usepackage{color}
\usepackage{hyperref}
\usepackage{algorithm}
\usepackage{algorithmic}
\usepackage{graphicx}
\usepackage{multirow, makecell}
\usepackage{caption}
\usepackage{afterpage}
\usepackage{pdflscape}
\usepackage{longtable}
\usepackage[table]{xcolor}
\usepackage{footnote}
\makesavenoteenv{tabular}
\makesavenoteenv{table}

\newcommand\numberthis{\addtocounter{equation}{1}\tag{\theequation}}
\DeclareMathOperator{\sign}{sgn}

\usepackage{xspace}
\newcommand*{\eg}{{\em e.g.}\@\xspace}
\newcommand*{\ie}{{\em i.e.}\@\xspace}
\newcommand*{\aka}{{\em a.k.a.}\@\xspace}
\newcommand*{\wrt}{{\em w.r.t.}\@\xspace}
\makeatletter
\newcommand*{\etc}{%
	\@ifnextchar{.}%
	{etc}%
	{etc.\@\xspace}%
}
\makeatother

\def\mybar#1{
	#1 & {\color{red}\rule{#1pt}{3pt}}}

\newcommand{\pc}{\cellcolor{purple!40}}
\newcommand{\rc}{\cellcolor{red!40}}
\newcommand{\bc}{\cellcolor{blue!40}}
\newcommand{\gc}{\cellcolor{green!40}}
\newcommand{\kc}{\cellcolor{black!40}}

\author{Tao Yang$^{1,2}$, Paul Thompson$^3$, Sihai Zhao$^4$ and Jieping Ye$^5$}
\title{Identifying Genetic Risk Factors via\\Sparse Group Lasso with Group Graph Structure}
\date{%
	\small
	$^1$Arizona State University, Tempe, AZ, USA \quad $^2$Baidu Research, Sunnyvale, CA, USA\\
	$^3$University of Southern California, Los Angeles, CA, USA \\
	$^4$University of Illinois at Urbana-Champaign, Champaign, IL, USA\\
	$^5$University of Michigan, Ann Arbor, MI, USA\\[1ex]
	{\normalsize\it t.yang@baidu.com, pthomp@usc.edu, sdzhao@illinois.edu, jpye@umich.edu}
}

\begin{document}
	
	\maketitle
	
	\begin{abstract}
		Genome-wide association studies (GWA studies or GWAS) investigate the relationships between genetic variants such as single-nucleotide polymorphisms (SNPs) and individual traits. 
		Recently, incorporating biological priors together with machine learning methods in GWA studies has attracted increasing attention. However, in real-world, nucleotide-level bio-priors have not been well-studied to date. Alternatively, studies at gene-level, for example, protein--protein interactions and pathways, are more rigorous and legitimate, and it is potentially beneficial to utilize such gene-level priors in GWAS. 
		In this paper, we proposed a novel two-level structured sparse model, called {\it Sparse Group Lasso with Group-level Graph structure} (SGLGG), for GWAS. It can be considered as a sparse group Lasso along with a group-level graph Lasso. Essentially, SGLGG penalizes the nucleotide-level sparsity as well as takes advantages of gene-level priors (both gene groups and networks), to identifying phenotype-associated risk SNPs. 
		We employ the alternating direction method of multipliers algorithm to optimize the proposed model. 
		Our experiments on the {\it Alzheimer's Disease Neuroimaging Initiative} whole genome sequence data and neuroimage data demonstrate the effectiveness of SGLGG. As a regression model, it is competitive to the state-of-the-arts sparse models; as a variable selection method, SGLGG is promising for identifying Alzheimer's disease-related risk SNPs.
	\end{abstract}
	
	\section{Introduction}
	Genetic variation is what makes us all unique. It refers to the diversity in the DNA sequence in human genomes and it may affect how an individual develops a disease or and responds to drugs, vaccines, pathogens, and etc \cite{carlson2008snps, blazer2006genes}. The most common type of genetic variation is a single-nucleotide polymorphism (SNP)---\ie, a difference in a single nucleotide in the deoxyribonucleic acid (DNA) \cite{kidd2008mapping}. In the past decade, genome-wide association studies (GWA studies or GWAS), which aim at revealing the relationships between genetic variants such as SNPs and individual traits, have attracted much attention achieved considerable success \cite{korte2013advantages, visscher2012five, welter2013nhgri}.
	
	Traditional GWA studies are based on statistical tests. Genetic risk factors are determined by their statistical significance, where a general procedure is to perform a statistical test between each individual SNP and the phenotype under investigation \cite{wray2013pitfalls, edwards2013beyond, clarke2011basic}. 
	For example, via meta-analyses, 11 new susceptibility SNPs for Alzheimer's disease (AD) have been identified \cite{lambert2013meta}; 10 loci that may influence allergic sensitization have been detected \cite{bonnelykke2013meta}.
	However, such kind of approaches has several limitations. First, it ignores the aggregate effects of multiple SNPs, for example, the epistatic interactions between loci \cite{zuk2012mystery, lippert2013exhaustive}. Second, independent SNP--phenotype testing disregards the SNPs' structural correlations associated with population genetics (\ie, linkage disequilibrium, LD) and biological relations (\eg functional relationships between genes) \cite{mieth2016combining}. 
	
	Later, increasing attention has been focused on Lasso (least absolute shrinkage and selection operator \cite{tibshirani1996regression}), as an alternative tool for identifying risk SNPs in GWAS \cite{yang2015detecting, Wang2016129}. Lasso is a multivariate method that models multiple SNPs simultaneously and highly precarious SNPs (that related to the phenotype under investigation) can be identified through the non-zero components of the model. For example, a previous whole genome association study \cite{yang2015detecting} shows Lasso together with stability selection \cite{meinshausen2010stability} is promising in detecting risk SNPs associated with Alzheimer's disease (AD) . However, there are two major drawbacks of Lasso: 1) it tends to arbitrary select only one from a set of highly correlated features \cite{hebiri2013correlations}; 2) it considers all features equally without any further structural assumptions. To address the above issues, utilizing structured sparse models together with different biological priors has aroused growing concern in GWAS, as incorporating such assumptions is favorable for model construction and interpretation \cite{ye2012sparse}. There are several attempts, for example, group Lasso \cite{liu2013incorporating}, tree Lasso \cite{Wang2016129}, and absolute fused Lasso \cite{yang2016absolute}. 
	
	It is worth mentioning that all those aforementioned approaches are based on the nucleotide-level biological assumptions (\eg LD or the consistency of successive SNPs). However, in real-world, at nucleotide-level, neither structural associations, nor functional relationships, nor interaction mechanisms, have been well-studied to date. On the other hand, studies of biological mechanisms are more rigorous and legitimate at gene-level. For example, GeneMANIA \cite{warde2010genemania} is a powerful tool for revealing gene-level biological networks. It integrates a large set of functional association data, including protein and genetic interactions, pathways, co-expression, co-localization and protein domain similarity. As a consequence, it is potentially beneficial to utilize such gene-level priors in nucleotide-level GWAS studies.
	
	In this paper, we propose a novel two-level structured sparse model, called {\it Sparse Group Lasso with Group-level Graph structure} (SGLGG), which a is promising method for identifying significant SNP--phenotype associations. As its name indicates, SGLGG can be considered as a fusion model of a sparse group Lasso \cite{Yuan2006, Friedman} and a group-level graph Lasso (\aka, graph-guided fused Lasso \cite{chen2010graph}). Essentially, our proposed model involves two levels of predictors---\ie, the nucleotide-level predictors and the gene-level predictors. And consequently in a GWA study, SGLGG will penalize the following three respects: 
	\begin{enumerate}\vspace{-0.5em}
		\setlength\itemsep{-0.3em}
		\item the gene-level sparsity;
		\item the graph structure among gene-level predictors;
		\item the nucleotide-level sparsity.
	\end{enumerate}
	\vspace{-0.5em}
	As a result, SGLGG tends to select only a set of causal SNPs within a gene group and limited gene groups among the entire sequence. Meanwhile, it is capable of taking advantages of biological priors (\ie, gene networks) during the gene-level selection. With the graph constraint, highly relevant genes are likely to be chosen simultaneously, and thus SNPs from different gene scopes are potentially able to connect. SGLGG is hard to solve due to its complex sparse-inducing regularizers. To this end, we first transfer the edge constraints among the graph into the matrix form, and then, employ the ADMM (alternating direction method of multipliers \cite{boyd2011distributed}) algorithm for optimizing. 
	Experiments have been conducted on the {\it Alzheimer's Disease Neuroimaging Initiative} (ADNI) whole genome sequence (WGS) data and neuroimage data, for both regression tasks and variable selection tasks. Preliminary results show that SGLGG is competitive to the state-of-the-arts sparse models in predicting AD-related imaging phenotypes. 
	In addition, stability selection results demonstrate that SGLGG is promising for identifying risk SNPs associated with Alzheimer's disease.	

%
%
%
%
	
	\section{Our Model: SGLGG}
	
	Essentially, we consider a linear prediction model. 
	Given a centered data matrix ${\bf A} \in \mathbb{R}^{n \times p}$ with $n$ observations and $p$ features, and a corresponding response ${\bf y} \in \mathbb{R}^n$. Suppose that $p$ predictors can be divided into $K$ non-overlapping groups, with $p_k$ the number of low-level predictors in group $k$. Accordingly, we denote ${\bf s} \in \mathbb{R}^p$ be the low-level predictors and ${\bf g} \in \mathbb{R}^K$ be the group-level predictors, respectively. Then, the low-level predictor ${\bf s}$ can be represented as ${\bf s} = [s_{11} \ldots s_{1p_1} ~\ldots~ s_{k1} \ldots s_{kp_k}]$. We further denote ${\bf G}_s = ({\bf M}^T{\bf g})\circ{\bf s} = [g_1s_{11} ~ g_1s_{12} \ldots g_1s_{1p_1} ~ g_2s_{21} ~ g_2s_{22} \ldots g_2s_{2p_2} ~\ldots~ g_ks_{kp_k}] \in \mathbb{R}^p$, where $\circ$ is the Hadamard product operator, ${\bf M} \in \mathbb{R}^{k \times p}$ is a designed mapping matrix\footnote{${\bf M} \in \mathbb{R}^{k \times p}$ is a binary matrix, an element $m_{ij}=1$ if $s_j$ in group $g_i$.}, and $g_i, i \in [1,k]$ is the $i$-th element of ${\bf g}$. 
	The group-level graph\footnote{In this study, we only consider the situation of undirected graph among group-level features.} information is described by $G\equiv(s_K,E)$, where $s_K=\{1,2,\ldots,k\}$ is a set of nodes, and $E$ is the set of edges. 
	In addition, let ${\bf w_g} \in \mathbb{R}^K$ denote the weight vector corresponding to the group-level predictors, and $r_{ij}$ denote the weight of the edge between node $g_i$ and $g_j$. Hence, in this paper, we consider the following optimization problem:
	\begin{align}\label{eq:prob}
	\min_{{\bf g}, {\bf s}}\bigg\{ 
	\ell({\bf y}, {\bf G}_s) + 
	\lambda_1 \| {\bf w_g \circ g} \|_1 + 
	\lambda_2 \sum_{(i,j)\in E} \tau(r_{ij}) {| g_i - \sign(r_{ij})g_j |} +
	\lambda_3 \| {\bf s} \|_1
	\bigg\},
	\end{align}
	where $\ell(\cdot)$ is a convex empirical loss function (\eg the least squares) and the error is calculated based on ${\bf G}_s$---a combination of predictors ${\bf g}$ and ${\bf s}$ via ${\bf M}$; $\lambda_1, \lambda_2, \lambda_3 \geq 0$; and $\tau(r_{ij})$ represent a general monotonically increasing function weight function that enforces a fusion effect between coefficients $g_i$ and $g_j$.
	
	In Eq. \eqref{eq:prob}, the first constraint can be considered as a group-level sparsity constraint, the second constraint introduces the group-level graph structure via the fused Lasso, and the third constraint penalizes the low-level sparsity. 
	Hereby, we call Problem \eqref{eq:prob}, the {\it Sparse Group Lasso with Group-level Graph structure} (SGLGG) problem. 
	More specifically, in a GAW study, ${\bf s}$ represents the nucleotide-level predictor, and accordingly, ${\bf g}$ can be considered as the gene-level predictor. Therefore, an ideal solution to Eq. \eqref{eq:prob} will lead to the following scenarios: 1) only limited gene groups will be selected among the entire sequence; 2) the group selection is guided by the gene-level biological priors---\ie, relevant genes are more likely to be chosen simultaneously; and 3) only a subset of SNPs will be selected within a selected gene. In other words, the gene-level and nucleotide-level constraints ensure that the most relevant gene groups and SNPs within a gene will be chosen by the model. Meanwhile, the group selection will be affected by the gene-level priors---\ie, some inter-gene SNP--SNP connections could be revealed by the graph constraint.
	
	Furthermore, the grpah constriant in Eq. \eqref{eq:prob} can be reformualted into a matrix form. Denote ${\bf T}$ be the sparse matrix constructed from the edge set $E$, where $t_{ij} = t_{ji} = r_{ij}$ if there is a edge between $g_i$ and $g_j$.
	Furthermore, for discussion convenience, we ignore the weight vectors in Eq. \eqref{eq:prob}, then SGLGG problem can be simplified as the following matrix form:
	\begin{align}\label{eq:prob_mat}
	\min_{{\bf g}, {\bf s}}{ 
		\ell({\bf y}, {\bf G}_s) + 
		\lambda_1 \| {\bf g} \|_1 + 
		\lambda_2 \| {\bf Tg} \|_1  +
		\lambda_3 \| {\bf s} \|_1
	}.
	\end{align}
	
	\section{ADMM for Solving SGLGG}
	
	\subsection{ADMM basic}
	
	Due to the complex sparse-inducing regularizers, unconstrianted optimzation problem like \eqref{eq:prob} are sometimes hard to solve directly. Instead, it is possbile to reformulate the original unconstrianted problem to an equivalent constrained problem. In the sequel, such a problem can be addressed using constrained optimization methods such as the augmented Lagrangian method. 
	
	Hereby, we employ the alternating direction method of multipliers (ADMM) \cite{boyd2011distributed, parikh2014proximal} algorithm to solve Problem \eqref{eq:prob}. ADMM is a variant of the augmented Lagrangian method. It utilizes dual decomposition and  partial updates for the dual variables.	Without loss of generality, we consider the following constraint optimization problem:
	\begin{align*}\label{eq:admm-basic}
	\min_{\bf{x,z}} { f({\bf x}) + g({\bf z}) }\numberthis\\
	\textit{~s.t.~} \bf{Ax+Bz=c},
	\end{align*}
	where $f$ and $g$ are convex, ${\bf x}\in\mathbb{R}^p$, ${\bf z}\in\mathbb{R}^q$, ${\bf A}\in \mathbb{R}^{n \times p}$, ${\bf B}\in \mathbb{R}^{n \times q}$, and ${\bf c}\in \mathbb{R}^{n}$. With ADMM, we first reformulate the above problem \eqref{eq:admm-basic} as:
	\begin{align}\label{admm-augmented}
	L_\rho({\bf{x,z,\mu}}) = f({\bf x}) + g({\bf z}) + {\bf \mu}^T({\bf Ax+Bz-c}) + \frac{\rho}{2}{\| \bf{Ax+Bz-c}  \|^2},
	\end{align}
	with ${\bf \mu}$ being the augmented Lagrangian multiplier, and $\rho$ being the non-negative dual update step length. ADMM solves this problem by iteratively minimizing $L_\rho(\bf{x,z,\mu})$ over ${\bf x}$, ${\bf z}$ and ${\bf \mu}$, one at a time, until convergence. Consequently, the update rule for ADMM is given by
	\begin{align*}
	{\bf x}^{k+1} & := \arg\min_{\bf x} L_\rho( {\bf x}, {\bf z}^k, {\bf \mu}^k ),\\
	{\bf z}^{k+1} & := \arg\min_{\bf z} L_\rho( {\bf x}^{k+1}, {\bf z}, {\bf \mu}^k ),\\
	{\bf \mu}^{k+1} & := {\bf \mu}^k + \rho( {\bf A}{\bf x}^{k+1} + {\bf B}{\bf z}^{k+1} - {\bf c} ).
	\end{align*}
	
	\subsection{ADMM for solving SGLGG problem}
	
	Suppose $\ell(\cdot)$ be the least squares loss, then the SGLGG problem presented in \eqref{eq:prob_mat} can be rewritten as the following constrained form:
	\begin{align*}\label{eq:prob_admm}
	\min_{{\bf g, s, p, q, r}} & { 
		\frac{1}{2}{ \| {\bf y} - {\bf A}{\bf G}_s \|^2 } + 
		\lambda_1 \| {\bf p} \|_1 + 
		\lambda_2 \| {\bf q} \|_1 +
		\lambda_3 \| {\bf r} \|_1
	} \numberthis\\
	\textit{s.t.~} & {\bf g-p=0}, {\bf Tg-q=0}, {\bf s-r=0},
	\end{align*}
	where ${\bf p},{\bf q},{\bf r}$ are slack variables. We employ ADMM to solve Problem \eqref{eq:prob_admm}. The augmented Lagrangian is
	\begin{align*}
	L_\rho(\bf{g, s, p, q, r, \mu, \nu, \xi}) = \numberthis \label{eq:lagrangian}
	\frac{1}{2}{ \| {\bf y} - {\bf A}{\bf G}_s \|^2 } + 
	& \lambda_1 \| {\bf p} \|_1 + 
	\lambda_2 \| {\bf q} \|_1 +
	\lambda_3 \| {\bf r} \|_1 +\\ 
	&{\bf \mu}^T({\bf g-p}) + 
	{\bf \nu}^T({\bf Tg-q}) + 
	{\bf \xi}^T({\bf s-r}) +\\
	&\frac{\rho}{2}\| {\bf g-p} \|^2 +
	\frac{\rho}{2}\| {\bf Tg-q} \|^2 +
	\frac{\rho}{2}\| {\bf s-r} \|^2,
	\end{align*}
	where $\bf{\mu, \nu, \xi}$ are augmented Lagrangian multipliers. Accordingly, in the $(k+1)$-th iteration, the update rules are as follows:
	
	\begin{itemize}
		\item \textbf{Update ${\bf g}$}: ${\bf g}^{k+1}$ can be updated by minimizing $L_\rho$ with ${\bf s, p, q, r, \mu, \nu, \xi}$ fixed:
		\begin{align*}
		{\bf g}^{k+1}
		=& \arg\min_{\bf g}{ \frac{1}{2}{ \| {\bf y} - {\bf A}[({\bf M}^T{\bf g})\circ{\bf s}^k] \|^2 } + 
			({\bf \mu}^k + {\bf T}^T{\bf \nu}^k)^T{\bf g} } +
		\frac{\rho}{2}\| {\bf g} - {\bf p}^k \|^2 + \frac{\rho}{2}\| {\bf Tg} - {\bf q}^k \|^2\\
		=& \arg\min_{\bf g}{ \frac{1}{2}{ \| {\bf y} - {\bf A}{\it Diag}({\bf s}^k){\bf M}^T{\bf g} \|^2 } + 
			[({\bf \mu}^k + {\bf T}^T{\bf \nu}^k)^T  - \rho {\bf p}^k - \rho {\bf T}^T{\bf q}^k ] {\bf g} } + \frac{\rho}{2}{ {\bf g}^T ({\bf I} + {\bf T}^T{\bf T}) {\bf g} }\\
		=& \arg\min_{\bf g}{ \frac{1}{2}{ {\bf g}^T[({\bf B}^k)^T{\bf B}^k + \rho({\bf I} + {\bf T}^T{\bf T}) ]{\bf g} 
				- [{\bf y}^T{\bf B}^k - ({\bf \mu}^k + {\bf T}^T{\bf \nu}^k)^T }} + \rho({\bf p}^k)^T + \rho({\bf q}^k)^T{\bf T} ] {\bf g}
		\end{align*}
		where ${\bf B}^k = {\bf A}{\it Diag}({\bf s}^k){\bf M}^T$, and ${\it Diag}(\cdot)$ is an operation that transforms a vector into a square diagonal matrix. The above optimization problem is quadratic, and thus the optimal solution can be obtained by solving the following linear system:
		\begin{align}\label{eq:g_update}
		{\bf F}_g^k{\bf g}^{k+1} = {\bf b}_g^k,
		\end{align}
		where
		\begin{align*}
		{\bf F}_g^k & = ({\bf B}^k)^T{\bf B}^k + \rho({\bf I} + {\bf T}^T{\bf T}),\\
		{\bf b}_g^k & = ({\bf B}^k)^T{\bf y} - {\bf \mu}^k - {\bf T}^T{\bf \nu}^k + \rho {\bf p}^k + \rho {\bf T}^T{\bf q}^k.
		\end{align*}
		It is trivial to show that ${\bf F}_g^k$ is {\it symmetric positive definite} (SPD), and thus Eq. \eqref{eq:g_update} can be solved efficiently via the conjugate gradient method \cite{hestenes1952methods} .
		
		\item \textbf{Update ${\bf s}$}: ${\bf s}^{k+1}$ can be updated by minimizing $L_\rho$ with ${\bf g, p, q, r, \mu, \nu, \xi}$ fixed:
		\begin{align*}
		{\bf s}^{k+1} 
		& = \arg\min_{\bf s}{ \frac{1}{2}{ \| {\bf y} - {\bf A}[({\bf M}^T{\bf g}^{k+1}) \circ {\bf s}] \|^2 } + 
			({\bf \xi}^k)^T{\bf s} + \frac{\rho}{2}\| {\bf s}-{\bf r}^k \|^2}\\
		& = \arg\min_{\bf s}{ \frac{1}{2}{ \| {\bf y} - {\bf A} {\it Diag}({\bf M}^T{\bf g}^{k+1}){\bf s} \|^2 } + 
			({\bf \xi}^k)^T{\bf s} + \frac{\rho}{2}\| {\bf s}-{\bf r}^k \|^2}\\
		& = \arg\min_{\bf s}{ \frac{1}{2}{ {\bf s}^T[({\bf C}^{k})^T{\bf C}^{k} + \rho{\bf I} ]{\bf s} -
				[{\bf y}^T{\bf C}^{k} - ({\bf \xi}^k)^T + \rho({\bf r}^k)^T] }{\bf s} },
		\end{align*}
		where ${\bf C}^{k} = {\bf A}{\it Diag}({\bf M}^T{\bf g}^{k+1})$. Similar to the update rule of ${\bf g}$, the above optimization problem is quadratic, and thus the optimal solution can be obtained by solving the following linear system:
		\begin{align}\label{eq:s_update}
		{\bf F}_s^k{\bf s}^{k+1} = {\bf b}_s^k,
		\end{align}
		where
		\begin{align*}
		{\bf F}_s^k & = {\bf C}^T{\bf C} + \rho{\bf I},\\
		{\bf b}_s^k & = {\bf C}^T{\bf y} - {\bf \xi}^k + \rho{\bf r}^k.
		\end{align*}
		Similarly, since ${\bf F}_s^k$ is SPD, Eq. \eqref{eq:s_update} can be solved efficiently via the conjugate gradient method.
		
		\item \textbf{Update ${\bf p}$}: Similarly, ${\bf p}^{k+1}$ can be obtained by solving the following problem:
		\begin{align*}
		{\bf p}^{k+1} 
		& = \arg\min_{\bf p}{ \lambda_1 \| {\bf p} \|_1 + ({\bf \mu}^k)^T({\bf g}^{k+1}-{\bf p}) + \frac{\rho}{2}\| {\bf g}^{k+1}-{\bf p} \|^2 }\\
		& = \arg\min_{\bf p}{ \lambda_1 \| {\bf p} \|_1 - ({\bf \mu}^k)^T{\bf p} + \frac{\rho}{2}\| {\bf g}^{k+1}-{\bf p} \|^2 }\\
		& = \arg\min_{\bf p}{ \frac{1}{2}\| {\bf p} - ({\bf g}^{k+1}+ \frac{1}{\rho}{{\bf \mu}^k}) \|^2 + \frac{\lambda_1}{\rho}{ \| {\bf p} \|_1 } }
		\end{align*}
		The above optimization problem has a closed-firm solution, known as the \textit{soft-thresholding}:
		\begin{align}\label{eq:p_update}
		{\bf p}^{k+1} = S_{\lambda_1/\rho}{ ({\bf g}^{k+1}+ \frac{1}{\rho}{{\bf \mu}^k}) },
		\end{align}
		where the \textit{soft-thresholding operator} is defined as:
		\begin{align*}
		S_{\lambda}(x) = \sign(x) \max{(|x|-\lambda, 0)}.
		\end{align*}
		
		\item \textbf{Update ${\bf q}$}: Similarly, ${\bf q}^{k+1}$ can be obtained by solving the following problem:
		\begin{align*}
		{\bf q}^{k+1} 
		& = \arg\min_{\bf q}{ \lambda_2 \| {\bf q} \|_1 + ({\bf \nu}^k)^T({\bf T}{\bf g}^{k+1}-{\bf q}) + \frac{\rho}{2}\| {\bf T}{\bf g}^{k+1}-{\bf q} \|^2 }.
		\end{align*}
		The closed-form solution of the above problem can be obtained by:
		\begin{align}\label{eq:q_update}
		{\bf q}^{k+1} = S_{\lambda_2/\rho}{ ({\bf T}{\bf g}^{k+1}+ \frac{1}{\rho}{{\bf \nu}^k}) }.
		\end{align}
		
		\item \textbf{Update ${\bf r}$}: Similarly, ${\bf r}^{k+1}$ can be obtained by solving the following problem:
		\begin{align*}
		{\bf r}^{k+1} 
		& = \arg\min_{\bf r}{ \lambda_3 \| {\bf r} \|_1 + ({\bf \xi}^k)^T({\bf s}^{k+1}-{\bf r}) + \frac{\rho}{2}\| {\bf s}^{k+1}-{\bf r} \|^2 }.
		\end{align*}
		The closed-form solution of the above problem can be obtained by:
		\begin{align}\label{eq:r_update}
		{\bf r}^{k+1} = S_{\lambda_3/\rho}{ ({\bf s}^{k+1}+ \frac{1}{\rho}{{\bf \xi}^k}) }.
		\end{align}
		
		\item \textbf{Update ${\bf \mu, \nu, \xi}$}: In the $(k+1)$-th iteration, ${\bf \mu, \nu, \xi}$ are updated by:
		\begin{align}
		{\bf \mu}^{k+1} & = {\bf \mu}^{k} + \rho( {\bf g}^{k+1} - {\bf p}^{k+1}), \label{eq:mu_update}\\
		{\bf \nu}^{k+1} & = {\bf \nu}^{k} +  \rho( {\bf T}{\bf g}^{k+1} - {\bf q}^{k+1}), \label{eq:nu_update}\\
		{\bf \xi}^{k+1} & = {\bf \xi}^{k} +  \rho( {\bf s}^{k+1} - {\bf r}^{k+1}). \label{eq:xi_update}
		\end{align}
	\end{itemize}
	
	We summarize the ADMM algorithm for solving the SGLGG Problem \eqref{eq:prob_mat} in Algorithm \ref{alg:admm}. 
	Generally, ADMM breaks the original complex optimization problem into a series of smaller subproblems, each of which is then easier to handle. In addition, it is worth mentioning that in practice, it is important to normalize $g_i$ according to its group size.
	
	\noindent\hspace{0.1\linewidth}\begin{minipage}{0.8\textwidth}
		\begin{algorithm}[H]
			\caption{ADMM for the sgLasso\_gGraph Problem}
			\label{alg:admm}
			\begin{algorithmic}[1]
				\REQUIRE{${\bf A}, {\bf y}, E, \lambda_1, \lambda_2, \lambda_3, \rho$\\}
				\ENSURE  ${\bf g, s}$\\
				\STATE{Initialization: Initialize ${\bf g}$ and ${\bf s}$, $k \leftarrow 0$.}
				\WHILE{not converge}
				\STATE{Update ${\bf g}^{k+1}$ according to Eq. \eqref{eq:g_update}.}
				\STATE{Update ${\bf s}^{k+1}$ according to Eq. \eqref{eq:s_update}.}
				\STATE{Update ${\bf p}^{k+1}$ according to Eq. \eqref{eq:p_update}.}
				\STATE{Update ${\bf q}^{k+1}$ according to Eq. \eqref{eq:q_update}.}
				\STATE{Update ${\bf r}^{k+1}$ according to Eq. \eqref{eq:r_update}.}
				\STATE{Update ${\bf \mu}^{k+1}, {\bf \nu}^{k+1} \text{~and~} {\bf \xi}^{k+1}$ according to Eqs, respectively. \eqref{eq:mu_update}, \eqref{eq:nu_update} \& \eqref{eq:xi_update}.}
				\ENDWHILE \\
			\end{algorithmic}
		\end{algorithm}
	\end{minipage}\\
		
	\section{Experiments}
	
	To evaluate the performance of the proposed SGLGG approach in GWAS, we conducted a series of experiments on the {\it Alzheimer's Disease Neuroimaging Initiative} (ADNI) whole genome sequence (WGS) data and neuroimage data. Particularly, we focus on two learning tasks: 1) predicting AD-related imaging phenotypes (based on SNPs data); and 2) identifying risk SNPs \wrt AD imaging phenotypes.
	
	\subsection{Data processing}
	
	\subsubsection{ADNI WGS data and neuroimaging data}
	
	In this study, we adopt the ADNI WGS data set and MRI data for GWAS. 
	More specifically, the following procedures have been employed for processing SNPs data. First, we employ PLINK \cite{purcell2007plink} together with a series of standard quality control constraints for SNPs data preprocessing. Particularly, a SNP will be removed if its minor allele frequency (MAF) $< 5\%$, or missingness $> 5\%$, or deviations from Hardy-Weinberg Equilibrium $P < 5 \times 10^{-7}$. In the sequel, we adopt MaCH \cite{li2010mach} for genotype imputation. MaCH is a Markov chain based haplotyper that is capable of resolving long haplotypes or inferring missing genotypes. Eventually, we apply several filters on the imputed data set, including: RSQ (estimated $R^2$, specific to each SNP) $> 0.5$, FREQ1 (frequency for reference Allele 1) $> 1\%$ and FREQ1 $< 99\%$. As a consequence, the entire genome data contains 1,319 subjects with 6,566,154 SNPs, in which 155,357 SNPs are from Chromosome 19. For subjects composition, there are 327 healthy controls (HC), 249 AD patients, 41 participants with mild cognitive impairment (MCI), 220 early MCI (EMCI) patients, 419 late MCI (LMCI) patients, and 63 patients with significant memory concerns (SMC). 
	
	Volumes of some major influenced brain regions that are related to Alzheimer's disease, including the hippocampus (HIPP) and the entorhinal cortex (EC), have been chosen as the neuroimaging phenotypes in this study. Those volumes were extracted from subject's T1 MRI data using Freesurfer \cite{reuter:long12},

	\subsubsection{Candidate AD genes}
	
\begin{figure}[hbpt!]
		\centering
		\includegraphics[width=4.2in]{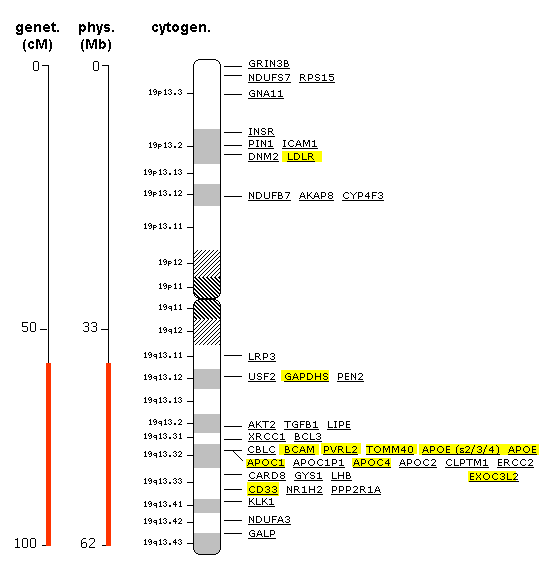}
		\captionsetup{format=hang,justification=centering}
		\caption{AD-risk genes (marked by yellow) on Chr.19 according to AlzGene. \\Figure adapted from: \url{http://www.alzgene.org/chromo.asp?c=19}}
		\label{fig:gene_pos}
	\end{figure}

	Hereby, we focus on Alzheimer's disease genetic risk factors (at both gene-level and nucleotide-level) on the 19th chromosome of the human genome. Particularly, at gene-level, ten candidate genes are pre-selected as high AD-risk according to AlzGene \cite{bertram2007systematic}, including {\it LDLR, GAPDHS, BCAM, PVRL2, TOMM40, APOE, APOC1, APOC4, EXOC3L2}, and {\it CD33}. Positions of those pre-selected genes are shown in Figure \ref{fig:gene_pos}.
	
	The above ten genes have been considered as the most strongly associated genes with AD on Chromosome 19 (Chr.19). In AlzGene, top associated genes are ranked based on genetic variants with the best overall HuGENet/Venice grades \cite{ioannidis2008assessment}. Specifically, for genes with identical grades, the ranking is based on their p-values; for genes with identical grades \& p-values, the ranking is based on their effect sizes. Basic information on those AD-risk genes is available in Table \ref{tab:gene_info} (top part).
	
	\subsubsection{Gene networks} \label{sec:gene_network}
	
\begin{figure}[hbpt!]
		\centering
		\includegraphics[width=6.8in]{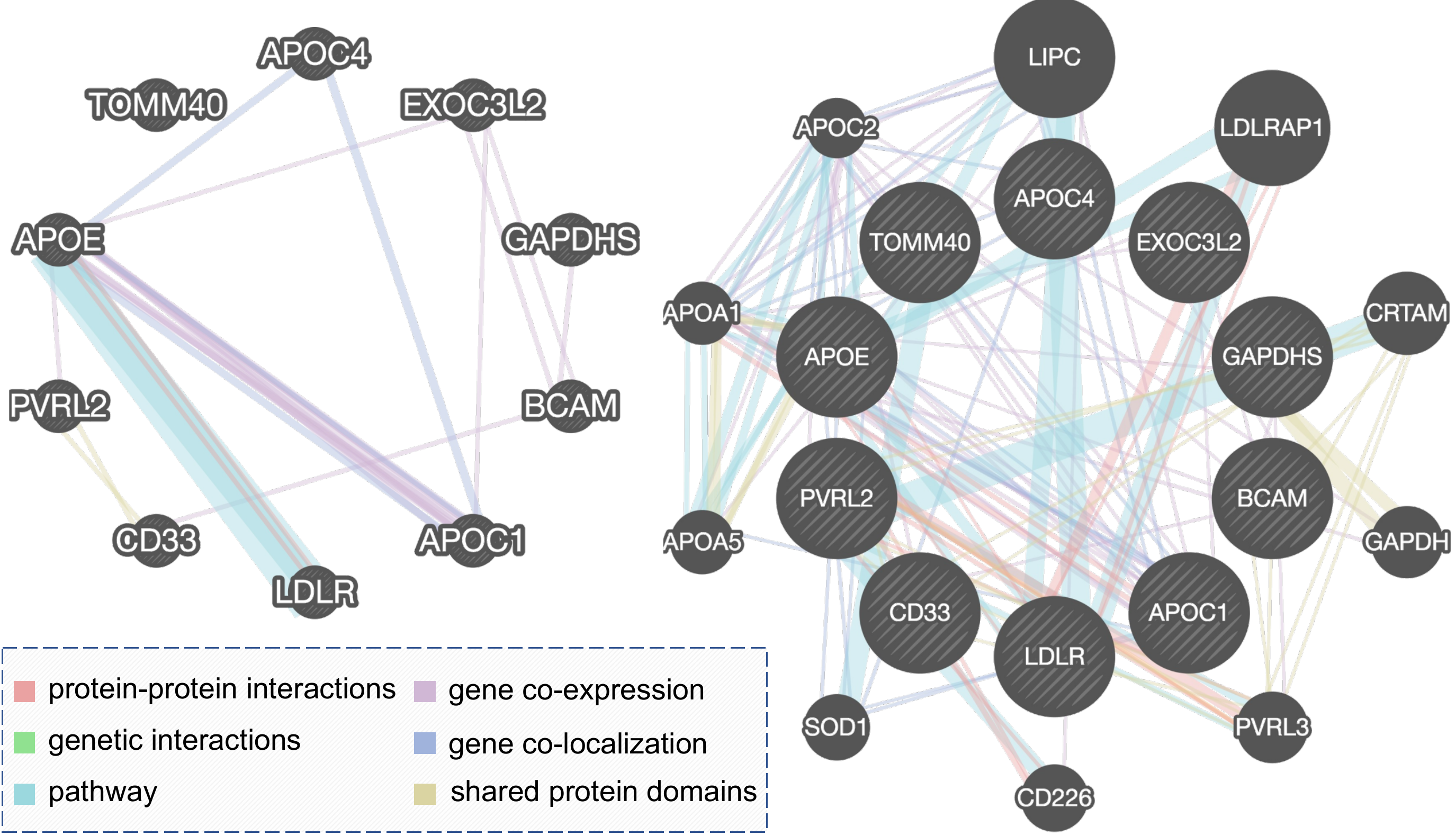}
		\caption{Visualizations of two gene networks. Left: network within 10 pre-selected AD-risk genes on Chr.19; Right: extended gene network based on 10 pre-selected Chr.19 AD-risk genes.}
		\label{fig:gene_networks}
	\end{figure}

	To retrieve gene-level biological priors---\ie, gene networks, we utilized GeneMANIA \cite{warde2010genemania} in our study. Essentially, GeneMANIA is a powerful tool to extract gene networks based on a set of input genes. The network is retrieved from a large set of functional association data, including gene co-expression \& co-localization, protein-protein interaction, genetic interaction, shared protein domains, pathway, and \etc. GeneMANIA stands for the {\em Multiple Association Network Integration Algorithm}. It consists of a linear regression-based algorithm for calculating the functional association network and a label propagation algorithm for predicting gene functions hereafter.	
	In our study, we employ the following two methods to extract gene networks.
	\begin{enumerate}
		\item {\bf Gene network within 10 pre-selected AD-risk genes in Chr.19}.\\
		Ten aforementioned AD-risk genes on Chromosome 19 are utilized as the input genes for GeneMANIA. For network exploration, we only focus on connections within those ten pre-selected genes. In addition, we adopt the biological process-based method for gene ontology weighting. A visualization of this gene networks is shown in Figure \ref{fig:gene_networks} (left). 
		\item {\bf Extended gene network based on 10 selected Chr19 AD-related genes}.\\
		Similar to 1, but we allow to introduce ten additional genes for network exploration. This results in totally 20 genes in the graph. A visualization of such a network is shown in Figure \ref{fig:gene_networks} (right). Note that, additional genes are selected based on their relations with input genes and thus those genes are not necessary located on Chromosome 19. Additional information of those selected genes is available in Table \ref{tab:gene_info} (bottom part).
	\end{enumerate}

	\begin{table}[t]
		\small
		\centering
		\caption{Basic information of selected genes}\vspace{2mm}
		\label{tab:gene_info}
		\def\arraystretch{1}
		\begin{tabular}{c|l|c|c|l|c}
			& \multicolumn{1}{c|}{\bf Symbol} & {\bf Assembly}   & {\bf Chr} & \multicolumn{1}{c|}{{\bf Location}} & {\bf \# of loci}\footnote{This is the number of available loci in our experimental dataset.} \\ \hline
			\parbox[t]{5mm}{\multirow{10}{*}{\rotatebox[origin=c]{90}{AD Candidate Genes}}}
			& LDLR                        & GRCh37.p13 & 19  & 11200037..11244506            & 135        \\ 
			& GAPDHS                      & GRCh37.p13 & 19  & 36024314..36036221            & 22         \\ 
			& BCAM                        & GRCh37.p13 & 19  & 45312316..45324678            & 15         \\ 
			& PVRL2                       & GRCh37.p13 & 19  & 45349393..45392485            & 164        \\ 
			& TOMM40                      & GRCh37.p13 & 19  & 45394477..45406946            & 38         \\
			& APOE                        & GRCh37.p13 & 19  & 45409039..45412650            & 5          \\ 
			& APOC1                       & GRCh37.p13 & 19  & 45417577..45422606            & 14         \\
			& APOC4                       & GRCh37.p13 & 19  & 45445495..45448753            & 7          \\ 
			& EXOC3L2                     & GRCh37.p13 & 19  & 45715879..45737469            & 88         \\  
			& CD33                        & GRCh37.p13 & 19  & 51728335..51743274            & 16         \\ \hline
			\parbox[t]{5mm}{\multirow{10}{*}{\rotatebox[origin=c]{90}{Associated Genes}}}
			& LDLRAP1                     & GRCh37.p13 & 1   & 25870071..25895377            & 28           \\ 
			& PVRL3                       & GRCh37.p13 & 3   & 110790606..110913017          & 73           \\ 
			& APOA5                       & GRCh37.p13 & 11  & 116660086..116663136          & 7           \\
			& APOA1                       & GRCh37.p13 & 11  & 116706467..116708338          & 5           \\ 
			& CRTAM                       & GRCh37.p13 & 11  & 122709255..122743347          & 75           \\
			& GAPDH                       & GRCh37.p13 & 12  & 6643585..6647537              & 10           \\
			& LIPC                        & GRCh37.p13 & 15  & 58702953..58861073            & 481           \\ 
			& CD226                       & GRCh37.p13 & 18  & 67530192..67624412            & 149           \\ 
			& APOC2                       & GRCh37.p13 & 19  & 45449239..45452822            & 17           \\  
			& SOD1                        & GRCh37.p13 & 21  & 33031935..33041244            & 15           \\ 
		\end{tabular}
	\end{table}
	
%

	Later, the experimental data sets were generated through those two aforementioned methods. More specifically, we first construct a smaller SNPs data set that consists of SNPs from 10 pre-selected AD-risk genes on Chromosome 19. As a result, such a data set contains 1,381 subjects and 504 SNPs. Next, we generate a larger SNPs data set based on an extended gene network obtained through GeneMANIA---\ie, SNPs from 10 additional genes (as shown in Table \ref{tab:gene_info}) are also involved, according to gene-level associations. Accordingly, the larger SNPs data set contains 1,364 SNPs in total from 20 candidate genes.
	
	\subsection{Learning task I --- Predicting AD-related phenotypes}
	
	In the first series of experiments, we evaluate our proposed SGLGG model in a set of regression tasks---\ie, predicting Alzheimer's disease-related imaging phenotypes. More specifically, SGLGG is compared with a suite of well-known commonly-used (structured) sparse methods, including Lasso, the fused Lasso (FL) and sparse group Lasso (SGL). For SGL and SGLGG, SNPs in the same gene naturally fall into a group. In addition, we compare SGLGG with the absolute fused Lasso (AFL) \cite{yang2016absolute}---a novel learning model that penalizes SNPs successive similarities. Four imaging phenotypes including volumes of the left entorhinal cortex (LEH), left hippocampus (LHP), right entorhinal cortex (REH), and right hippocampus (RHP), are used as the responses in this study.
	
\begin{figure*}[hbpt!]
		\centering
		\includegraphics[width=5in]{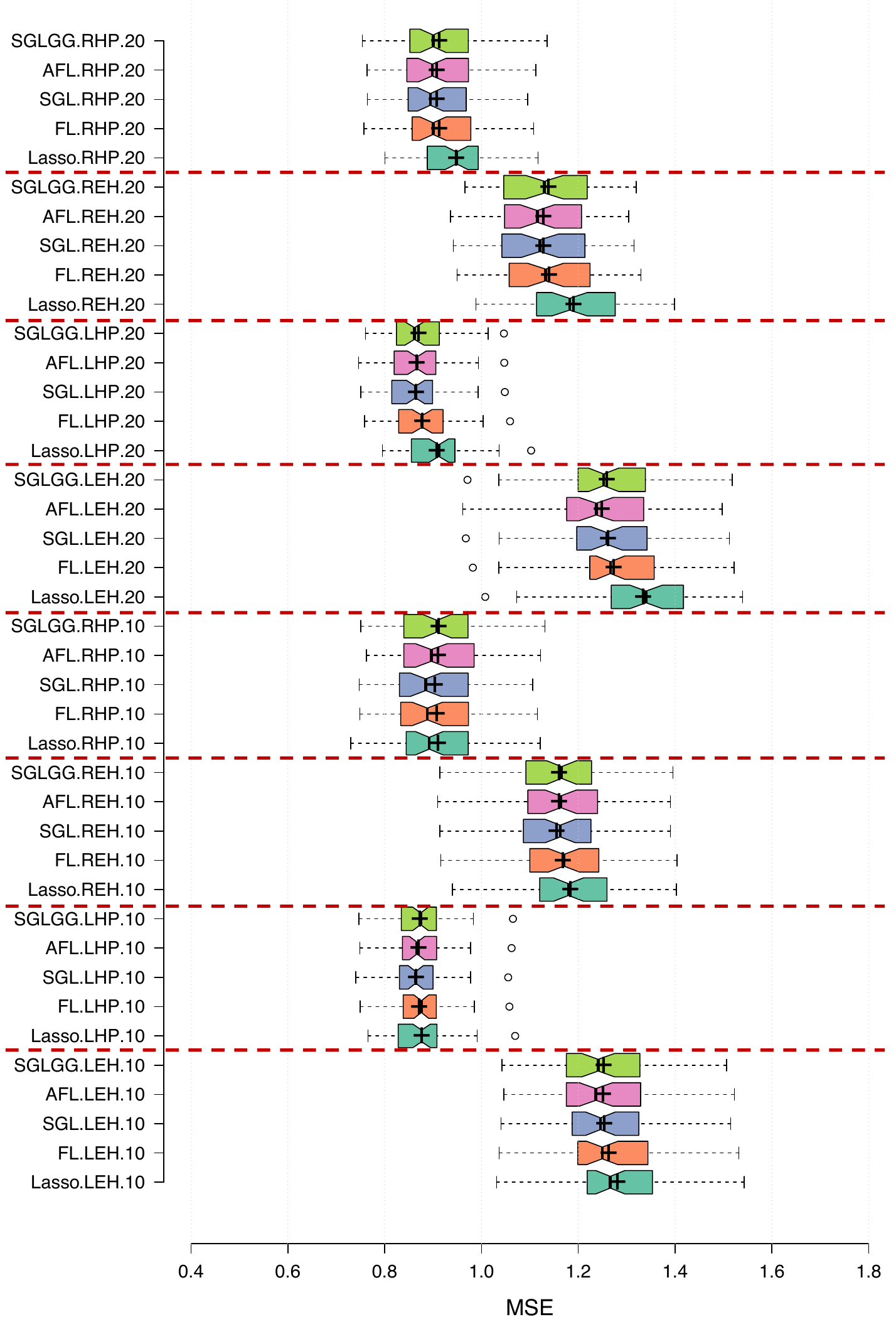}
		\captionsetup{format=hang}
		\caption{Comparison of regression error in terms of MSE of different structured sparse models on candidate AD-risk genes on Chr.19. For $y$-axis labels: the first few letters represent a modeling method, the middle three letters indicate the learning task, and the last number (10 or 20) indicate the data set involved.}
		\label{fig:adni_reg_comp}
	\end{figure*}

	Experiments have been conducted on the two SNPs data sets described in Section \ref{sec:gene_network}. We adopt five-fold cross-validation for each learning task and each sparse model. Comparisons of predictive performance in terms of mean squared error (MSE) of 10 replications are shown in Figure \ref{fig:adni_reg_comp} through box plots. In Figure \ref{fig:adni_reg_comp}, each color represents a modeling method. Labels of the $y$-axis are named as follows: the first few letters represent a modeling method, the middle three letters indicate the learning task, and the last number (10 or 20) indicate the data set involved.
	
	From Figure \ref{fig:adni_reg_comp}, we can observe that our proposed SGLGG model is very competitive compared with other (structured) sparse models. With complex sparse-inducing regularizers and complex bio-priors, SGLGG can still provide favorable predictive performance in most of the cases. Meanwhile, such a model has better interpretability than traditional ones, as it incorporated extensive prior knowledge during model learning. Therefore, it is potentially beneficial to address real-world GWA studies through the SGLGG model.
	
	\subsection{Learning task II --- Identifying AD-risk SNPs}
	
	One of the benefits of adopting a sparse model for GWAS is that the most relevant genetic factors can be identified through the non-zero components from the model. Hereby, in the following series of experiments, we compare the variable selection (\ie, SNPs selection) results of different structured sparse methods through stability selection \cite{meinshausen2010stability}. More specifically, experiments were conducted on the smaller SNPs data set mentioned in Sec \ref{sec:gene_network}. We perform 100 simulations for each learning target. Within each simulation, we first randomly subsample half of the subjects and then perform a modeling method 100 times with different regularization parameters (or pairs of parameters). The model selection results are visualized in Figure \ref{fig:adni_fs_comp}. Detailed SNPs selection results are available in Appendix \hyperref[sec:appendix]{1}.
	In Figure \ref{fig:adni_fs_comp}, top 50 selected SNPs are marked for each method; each color refers to a modeling method; the $x$-axis is a compact illustration of gene/SNPs location on Chromosome 19; green bars together with the $y$-axis indicate the negative logarithmic of P-values of SNPs associated with each learning task.
	
	From Figure \ref{fig:adni_fs_comp}, we have the following observations:
	\begin{enumerate}\vspace{-0.5em}
		\setlength\itemsep{-0.3em}
		\item SNPs selected by Lasso and SGL are spread over a large region in the feature sets (\ie, across different genes). However, most SNPs selected by FL, AFL, and our proposed SGLGG model are clustered in a few small regions.
		\item SNPs groups identified by SGLGG are different from FL or AFL, where the proposed method tends to select more SNPs within a gene but fewer number of genes in total.
		\item Statistical significance in terms of P-value of an SNP selected by SGLGG, may not necessarily be small\footnote{A smaller P-value implies higher statistical significance. Since we use the negative logarithm of P-values in Figure \ref{fig:adni_fs_comp}, statistically significant SNPs will have higher green bars.} (see the bottom two sub-figures in Figure \ref{fig:adni_fs_comp}). 
	\end{enumerate}
	
	The above observations imply that our proposed SGLGG model sparse selection on both nucleotide-level and gene-level. Within a gene, only the most relevant SNPs will be chosen. The group selection is benefited from gene-level biological prior knowledge---\ie, gene network. Thus, potential inter-gene SNP--SNP connections could be established by SGLGG. In other words, SGLGG is a promising method and has good prospects in revealing the causal SNPs that associated with a phenotype under investigation.
	
	\section{Conclusion}
	
	In this paper, we proposed a novel two-level structured sparse model---SGLGG---for genome-wide association studies. Essentially, it can be considered as a sparse group Lasso together with a group-level graph-guided fused Lasso. Specifically, SGLGG induces sparsities in both nucleotide-level and gene-level. That is, only the most causal SNPs will be selected within a gene group and only a part of relevant genes will be chosen on the genome. Another benefit of SGLGG is that it also takes advantages of gene-level biological priors during the model construction. Consequently, gene-level bio-priors such as protein--protein interactions and pathways can be utilized to explore inter-gene SNP--SNP connections. To address SGLGG model, we propose an ADMM-based optimization algorithm.
	Our experiments on the Alzheimer's disease genome sequence data and neuroimaging data show that SGLGG is very competitive in predict AD-related phenotypes, compared with other state-of-the-arts sparse learning models. Furthermore, stability selection results demonstrate that SGLGG is a promising model for identifying AD-risk SNPs. With the help of gene-level biological priors, SGLGG has good prospects for revealing SNP--SNP interactions among different genes.
	
	\section*{Acknowledgement}
	
	This work was supported in part by NIH BD2K (Big Data to Knowledge) grants to the KnowENG Center, based at UIUC, and the ENIGMA Center for Worldwide Medicine, Imaging \& Genomics, based at USC.
	
	{
		\bibliographystyle{abbrv}
		\bibliography{ref}}
	
	\afterpage{%
		\clearpage
		\begin{landscape}
			\centering 
			\begin{figure*}[!t]
				\centering
				\includegraphics[width=8.5in]{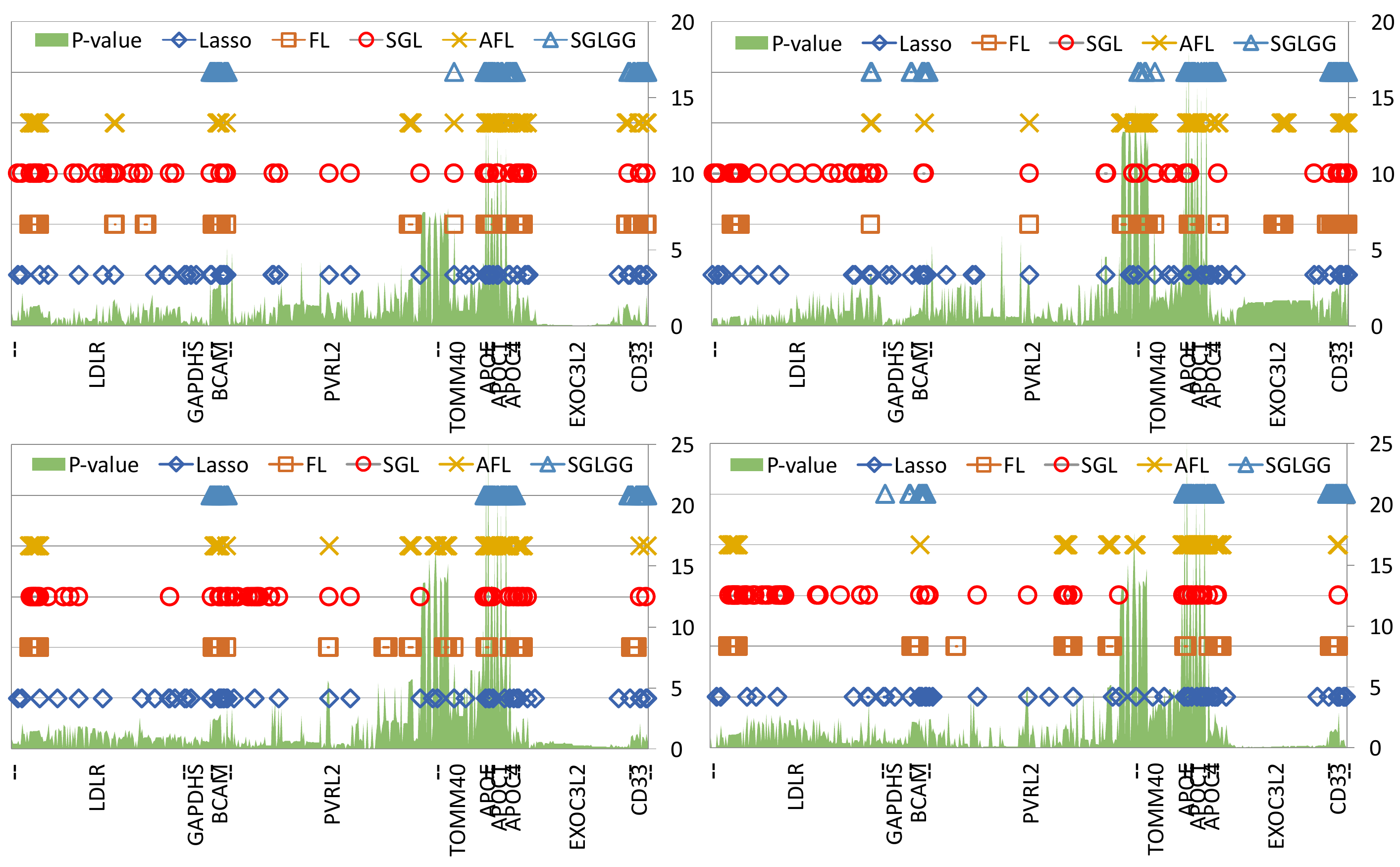}
				\captionsetup{format=hang}
				\caption{Comparison of stability selection results of different structured sparse models on Chromosome 19. Each subfigure refers to an AD-related neuroimaging phenotype, specifically, upper left---LEH; bottom left---LHP; upper right---REH; bottom right---RHP. The $x$-axis is a compact illustration of gene/SNP positions on Chr.19. The $y$-axis is the negative logarithm of P-value of SNPs regarding the phenotype under investigation. For each learning task, top 50 selected SNPs of each model are marked out.}
				\label{fig:adni_fs_comp}
			\end{figure*}
		\end{landscape}
		\clearpage
	}

	\appendix
	\renewcommand{\thesection}{\arabic{section}}

\afterpage{%
	\clearpage
	\begin{landscape}
		\section*{Appendix 1: Variable selection results of different learning targets} \label{sec:appendix}
		{	
			\footnotesize
			\centering 
			\setlength{\tabcolsep}{1.2pt}
			\renewcommand\arraystretch{0.78}
			\begin{longtable}{ll|lllll|ll|lllll|ll|lllll|ll|lllll|ll}
				\caption{Comparison of top 50 selected SNPs associated with different sparse models and neuroimage targets on the 10 AD-related genes on Chr.19. SNPs are sorted by their positions on the chromosome. Selection results are marked by different colors for different models. The negative logarithm of P-values is shown on the right side of each response. SNPs are ignored if they are not in the top selected list of any traits.\\ Model reference: L---Lasso; fL---fused Lasso; sgL---sparse group Lasso; afL---absolute fused Lasso; our---SGLGG. } \\
				& \multicolumn{1}{c|}{Target} & \multicolumn{7}{c|}{LEH} & \multicolumn{7}{c|}{LHP} & \multicolumn{7}{c|}{REH} & \multicolumn{7}{c}{RHP} \\ \cline{2-30} 
				Gene & \multicolumn{1}{l|}{Locus/SNP} & L & ~fL~ & sgL & afL & \multicolumn{1}{l|}{our} & \multicolumn{2}{c|}{-log(P)} & L & ~fL~ & sgL & afL & \multicolumn{1}{l|}{our} & \multicolumn{2}{c|}{-log(P)} & L & ~fL~ & sgL & afL & \multicolumn{1}{l|}{our} & \multicolumn{2}{c|}{-log(P)} & L & ~fL~ & sgL & afL & \multicolumn{1}{l|}{our} & \multicolumn{2}{c}{-log(P)} \\ \hline
				\endfirsthead
				\label{tab:fs_detailed}
				& \multicolumn{1}{c|}{Target} & \multicolumn{7}{c|}{LEH} & \multicolumn{7}{c|}{LHP} & \multicolumn{7}{c|}{REH} & \multicolumn{7}{c}{RHP} \\ \cline{2-30} 
				Gene & \multicolumn{1}{l|}{Locus/SNP} & L & ~fL~ & sgL & afL & \multicolumn{1}{l|}{our} & \multicolumn{2}{c|}{-log(P)} & L & ~fL~ & sgL & afL & \multicolumn{1}{l|}{our} & \multicolumn{2}{c|}{-log(P)} & L & ~fL~ & sgL & afL & \multicolumn{1}{l|}{our} & \multicolumn{2}{c|}{-log(P)} & L & ~fL~ & sgL & afL & \multicolumn{1}{l|}{our} & \multicolumn{2}{c}{-log(P)} \\ \hline
				\endhead
				\multirowcell{58}{LDLR\\(58/135)}    & rs12981050  &     &     &     &     &     & \mybar{0.7899}  &     &     &     &     &     & \mybar{0.7739}  & \kc &     &     &     &     & \mybar{0.1724}  &     &     &     &     &     & \mybar{0.8356}  \\
				& rs57217136  &     &     &     &     &     & \mybar{0.3490}  &     &     &     &     &     & \mybar{0.4339}  &     &     & \gc &     &     & \mybar{0.5000}  &     &     &     &     &     & \mybar{0.0073}  \\
				& 19:11201988 &     &     &     &     &     & \mybar{0.3486}  &     &     &     &     &     & \mybar{0.4326}  &     &     & \gc &     &     & \mybar{0.5004}  &     &     &     &     &     & \mybar{0.0066}  \\
				& 19:11202194 &     &     &     &     &     & \mybar{0.3415}  &     &     &     &     &     & \mybar{0.4172}  &     &     & \gc &     &     & \mybar{0.5083}  &     &     &     &     &     & \mybar{0.0007}  \\
				& rs6511720   &     &     &     &     &     & \mybar{0.3376}  &     &     &     &     &     & \mybar{0.4080}  & \kc &     & \gc &     &     & \mybar{0.5127}  &     &     &     &     &     & \mybar{0.0053}  \\
				& rs17242367  & \kc &     & \gc &     &     & \mybar{1.0509}  & \kc &     &     &     &     & \mybar{0.6456}  & \kc &     & \gc &     &     & \mybar{0.0168}  & \kc &     &     &     &     & \mybar{1.1834}  \\
				& rs6511721   & \kc &     & \gc &     &     & \mybar{2.1718}  & \kc &     &     &     &     & \mybar{2.0580}  & \kc &     &     &     &     & \mybar{0.5733}  & \kc &     &     &     &     & \mybar{1.8577}  \\
				& rs17248769  &     & \pc &     & \rc &     & \mybar{1.2221}  &     & \pc &     & \rc &     & \mybar{1.4125}  &     &     &     &     &     & \mybar{0.5339}  &     & \pc &     & \rc &     & \mybar{1.0289}  \\
				& rs73015033  &     & \pc & \gc & \rc &     & \mybar{1.2915}  &     & \pc & \gc & \rc &     & \mybar{1.4698}  &     & \pc & \gc &     &     & \mybar{0.5982}  &     & \pc & \gc & \rc &     & \mybar{1.0884}  \\
				& 19:11209669 &     & \pc &     & \rc &     & \mybar{1.1741}  &     & \pc & \gc & \rc &     & \mybar{1.4337}  &     & \pc & \gc &     &     & \mybar{0.5822}  &     & \pc &     & \rc &     & \mybar{1.0432}  \\
				& rs73015034  &     & \pc & \gc & \rc &     & \mybar{1.2951}  &     & \pc & \gc & \rc &     & \mybar{1.4728}  &     & \pc & \gc &     &     & \mybar{0.6028}  &     & \pc & \gc & \rc &     & \mybar{1.0910}  \\
				& 19:11209757 &     & \pc & \gc & \rc &     & \mybar{1.2972}  &     & \pc & \gc & \rc &     & \mybar{1.4749}  &     & \pc & \gc &     &     & \mybar{0.6051}  &     & \pc & \gc & \rc &     & \mybar{1.0929}  \\
				& rs74857287  &     & \pc & \gc & \rc &     & \mybar{1.2992}  &     & \pc & \gc & \rc &     & \mybar{1.4768}  &     & \pc & \gc &     &     & \mybar{0.6068}  &     & \pc & \gc & \rc &     & \mybar{1.0946}  \\
				& rs17248776  &     & \pc & \gc & \rc &     & \mybar{1.3039}  &     & \pc & \gc & \rc &     & \mybar{1.4794}  &     & \pc & \gc &     &     & \mybar{0.6098}  &     & \pc & \gc & \rc &     & \mybar{1.0980}  \\
				& rs17248783  &     & \pc & \gc & \rc &     & \mybar{1.3586}  &     & \pc & \gc & \rc &     & \mybar{1.5249}  &     & \pc & \gc &     &     & \mybar{0.6629}  &     & \pc & \gc & \rc &     & \mybar{1.1474}  \\
				& rs2228671   & \kc & \pc & \gc & \rc &     & \mybar{1.3720}  & \kc & \pc & \gc & \rc &     & \mybar{1.5250}  & \kc & \pc & \gc &     &     & \mybar{0.6681}  &     & \pc & \gc & \rc &     & \mybar{1.1210}  \\
				& rs36096887  &     &     &     &     &     & \mybar{0.5588}  &     &     &     &     &     & \mybar{0.6581}  &     &     &     &     &     & \mybar{0.5592}  &     &     & \gc &     &     & \mybar{1.7026}  \\
				& rs12983082  &     &     &     &     &     & \mybar{0.4876}  &     &     &     &     &     & \mybar{1.3053}  &     &     &     &     &     & \mybar{0.0132}  &     &     & \gc &     &     & \mybar{1.8414}  \\
				& rs10422256  & \kc &     & \gc &     &     & \mybar{0.9677}  &     &     & \gc &     &     & \mybar{2.0577}  &     &     &     &     &     & \mybar{0.0811}  & \kc &     & \gc &     &     & \mybar{2.7542}  \\
				& rs892116    &     &     &     &     &     & \mybar{0.3880}  &     &     &     &     &     & \mybar{1.4702}  &     &     &     &     &     & \mybar{0.4698}  &     &     & \gc &     &     & \mybar{2.4707}  \\
				& rs12710260  &     &     &     &     &     & \mybar{0.5528}  &     &     &     &     &     & \mybar{1.5892}  &     &     &     &     &     & \mybar{0.0134}  &     &     & \gc &     &     & \mybar{1.8819}  \\
				& 19:11223683 &     &     &     &     &     & \mybar{0.0027}  & \kc &     &     &     &     & \mybar{0.1174}  & \kc &     & \gc &     &     & \mybar{1.4241}  & \kc &     &     &     &     & \mybar{0.0745}  \\
				& rs9789328   &     &     &     &     &     & \mybar{0.5740}  &     &     & \gc &     &     & \mybar{1.9552}  &     &     &     &     &     & \mybar{0.0467}  &     &     & \gc &     &     & \mybar{2.1195}  \\
				& rs2738445   &     &     &     &     &     & \mybar{0.5148}  &     &     &     &     &     & \mybar{1.8265}  &     &     &     &     &     & \mybar{0.5569}  &     &     & \gc &     &     & \mybar{2.7873}  \\
				& rs2738446   &     &     &     &     &     & \mybar{0.5362}  &     &     & \gc &     &     & \mybar{1.9059}  &     &     &     &     &     & \mybar{0.0649}  &     &     & \gc &     &     & \mybar{2.0939}  \\
				& rs1799898   &     &     & \gc &     &     & \mybar{1.1926}  &     &     &     &     &     & \mybar{0.7147}  &     &     &     &     &     & \mybar{0.5502}  &     &     &     &     &     & \mybar{1.9748}  \\
				& rs28786710  &     &     &     &     &     & \mybar{0.5551}  &     &     &     &     &     & \mybar{1.8321}  &     &     &     &     &     & \mybar{0.1645}  &     &     & \gc &     &     & \mybar{2.0467}  \\
				& 19:11228620 & \kc &     & \gc &     &     & \mybar{0.9978}  & \kc &     & \gc &     &     & \mybar{1.8387}  & \kc &     & \gc &     &     & \mybar{1.1850}  & \kc &     &     &     &     & \mybar{0.0004}  \\
				& rs2738448   &     &     &     &     &     & \mybar{0.5506}  &     &     &     &     &     & \mybar{1.8251}  &     &     &     &     &     & \mybar{0.1691}  &     &     & \gc &     &     & \mybar{2.0366}  \\
				& rs2569550   &     &     &     &     &     & \mybar{0.1932}  &     &     &     &     &     & \mybar{1.5543}  &     &     &     &     &     & \mybar{0.7497}  &     &     & \gc &     &     & \mybar{2.6723}  \\
				& rs2738449   &     &     &     &     &     & \mybar{0.5585}  &     &     &     &     &     & \mybar{1.8219}  &     &     &     &     &     & \mybar{0.1763}  &     &     & \gc &     &     & \mybar{2.0355}  \\
				& rs2738450   &     &     &     &     &     & \mybar{0.5440}  &     &     &     &     &     & \mybar{1.8163}  &     &     &     &     &     & \mybar{0.1751}  &     &     & \gc &     &     & \mybar{2.0242}  \\
				& rs2738452   &     &     &     &     &     & \mybar{0.5423}  &     &     &     &     &     & \mybar{1.8135}  &     &     &     &     &     & \mybar{0.1769}  &     &     & \gc &     &     & \mybar{2.0201}  \\
				& rs12611153  &     &     & \gc &     &     & \mybar{1.4321}  &     &     &     &     &     & \mybar{0.6694}  &     &     & \gc &     &     & \mybar{2.0399}  &     &     &     &     &     & \mybar{0.7019}  \\
				& rs2569546   & \kc &     & \gc &     &     & \mybar{1.3039}  & \kc &     &     &     &     & \mybar{0.2799}  &     &     &     &     &     & \mybar{0.9068}  &     &     &     &     &     & \mybar{0.2749}  \\
				& rs2738454   &     &     & \gc &     &     & \mybar{1.3720}  &     &     &     &     &     & \mybar{0.7667}  &     &     &     &     &     & \mybar{2.0649}  &     &     &     &     &     & \mybar{0.6117}  \\
				& rs2738455   &     &     & \gc &     &     & \mybar{1.5490}  &     &     &     &     &     & \mybar{0.7928}  &     &     & \gc &     &     & \mybar{2.2572}  &     &     &     &     &     & \mybar{0.4317}  \\
				& 19:11235247 & \kc & \pc & \gc & \rc &     & \mybar{1.7311}  &     &     &     &     &     & \mybar{1.4309}  &     &     &     &     &     & \mybar{0.6046}  &     &     &     &     &     & \mybar{2.3629}  \\
				& rs8106324   &     & \pc &     & \rc &     & \mybar{1.1772}  &     &     &     &     &     & \mybar{0.9370}  &     &     &     &     &     & \mybar{0.2875}  &     &     &     &     &     & \mybar{1.9936}  \\
				& rs6511724   &     &     & \gc &     &     & \mybar{1.5654}  &     &     &     &     &     & \mybar{0.0920}  &     &     &     &     &     & \mybar{1.9689}  &     &     &     &     &     & \mybar{0.2419}  \\
				& rs75090161  &     &     &     &     &     & \mybar{0.6922}  &     &     &     &     &     & \mybar{0.7608}  &     &     &     &     &     & \mybar{0.4939}  &     &     & \gc &     &     & \mybar{2.2911}  \\
				& rs17242586  &     &     &     &     &     & \mybar{0.6864}  &     &     &     &     &     & \mybar{0.7536}  &     &     &     &     &     & \mybar{0.4904}  &     &     & \gc &     &     & \mybar{2.2829}  \\
				& rs2738457   &     &     & \gc &     &     & \mybar{1.5639}  &     &     &     &     &     & \mybar{0.7990}  &     &     & \gc &     &     & \mybar{2.2330}  &     &     &     &     &     & \mybar{0.4223}  \\
				& rs2569539   &     &     & \gc &     &     & \mybar{1.5293}  &     &     &     &     &     & \mybar{0.8141}  &     &     & \gc &     &     & \mybar{2.2363}  &     &     &     &     &     & \mybar{0.4221}  \\
				& rs2569538   &     &     &     &     &     & \mybar{0.7753}  & \kc &     &     &     &     & \mybar{1.6353}  &     &     &     &     &     & \mybar{0.4408}  &     &     & \gc &     &     & \mybar{2.1744}  \\
				& rs2738460   &     &     & \gc &     &     & \mybar{1.5121}  &     &     &     &     &     & \mybar{0.8278}  &     &     &     &     &     & \mybar{2.2257}  &     &     &     &     &     & \mybar{0.4269}  \\
				& rs2304182   &     & \pc &     &     &     & \mybar{1.0312}  &     &     &     &     &     & \mybar{0.9552}  &     &     &     &     &     & \mybar{0.9917}  &     &     &     &     &     & \mybar{0.4033}  \\
				& rs2304181   &     & \pc &     &     &     & \mybar{1.0317}  &     &     &     &     &     & \mybar{0.9557}  &     &     &     &     &     & \mybar{0.9915}  &     &     &     &     &     & \mybar{0.4038}  \\
				& 19:11239179 &     & \pc &     &     &     & \mybar{1.0319}  &     &     &     &     &     & \mybar{0.9558}  &     &     &     &     &     & \mybar{0.9918}  &     &     &     &     &     & \mybar{0.4039}  \\
				& rs2738461   &     &     &     &     &     & \mybar{2.0453}  &     &     &     &     &     & \mybar{0.3906}  & \kc &     & \gc &     &     & \mybar{2.7303}  &     &     &     &     &     & \mybar{0.0079}  \\
				& 19:11241428 & \kc &     &     &     &     & \mybar{0.2734}  & \kc &     &     &     &     & \mybar{0.6702}  & \kc &     & \gc &     &     & \mybar{0.4377}  & \kc &     &     &     &     & \mybar{0.3168}  \\
				& rs6413504   &     &     &     &     &     & \mybar{0.3292}  &     &     &     &     &     & \mybar{0.9297}  &     &     & \gc &     &     & \mybar{2.5425}  &     &     &     &     &     & \mybar{0.4233}  \\
				& rs2738464   &     &     &     &     &     & \mybar{0.2217}  &     &     &     &     &     & \mybar{0.6842}  &     &     &     &     &     & \mybar{1.1857}  &     &     & \gc &     &     & \mybar{2.1186}  \\
				& 19:11243209 & \kc &     &     &     &     & \mybar{0.6225}  & \kc &     &     &     &     & \mybar{0.9295}  & \kc &     & \gc &     &     & \mybar{0.2573}  & \kc &     &     &     &     & \mybar{0.8486}  \\
				& rs2915966   & \kc &     & \gc &     &     & \mybar{1.0703}  & \kc &     & \gc &     &     & \mybar{1.2712}  & \kc & \pc & \gc & \rc & \bc & \mybar{3.3284}  & \kc &     & \gc &     &     & \mybar{3.5090}  \\
				& rs2978615   &     &     &     &     &     & \mybar{0.7032}  &     &     &     &     &     & \mybar{0.6522}  & \kc &     & \gc & \rc & \bc & \mybar{3.5269}  &     &     &     &     &     & \mybar{0.1348}  \\
				& rs55903358  &     &     & \gc &     &     & \mybar{1.1494}  & \kc &     &     &     &     & \mybar{1.2808}  &     &     &     &     &     & \mybar{1.1002}  &     &     &     &     &     & \mybar{0.7609}  \\
				& rs5742911   &     &     &     &     &     & \mybar{1.1742}  &     &     &     &     &     & \mybar{0.9523}  &     &     & \gc &     &     & \mybar{2.0696}  &     &     &     &     &     & \mybar{0.5010}  \\ \hline
				\multirowcell{4}{GAPDHS\\(4/22)}  & rs4806174   & \kc &     &     &     &     & \mybar{0.2598}  & \kc &     &     &     &     & \mybar{1.0513}  &     &     &     &     &     & \mybar{0.2754}  & \kc &     &     &     &     & \mybar{1.1987}  \\
				& 19:36025093 & \kc &     &     &     &     & \mybar{0.2992}  & \kc &     &     &     &     & \mybar{0.4710}  & \kc &     &     &     &     & \mybar{0.2326}  & \kc &     &     &     & \bc & \mybar{1.6105}  \\
				& rs56408696  & \kc &     &     &     &     & \mybar{0.9017}  & \kc &     &     &     &     & \mybar{0.6562}  & \kc &     &     &     &     & \mybar{0.4423}  &     &     &     &     &     & \mybar{0.3527}  \\
				& rs2239942   & \kc &     &     &     &     & \mybar{0.5671}  &     &     &     &     &     & \mybar{0.6892}  &     &     &     &     &     & \mybar{0.4964}  & \kc &     &     &     &     & \mybar{0.3936}  \\ \hline
				\multirowcell{15}{BCAM\\(15/15)}    & rs2927477   & \kc &     & \gc &     & \bc & \mybar{2.9703}  & \kc &     &     &     & \bc & \mybar{2.4167}  &     &     &     &     & \bc & \mybar{1.3773}  &     &     &     &     & \bc & \mybar{1.5632}  \\
				& 19:45314324 & \kc &     &     &     & \bc & \mybar{0.0843}  & \kc &     & \gc &     & \bc & \mybar{0.6623}  & \kc &     &     &     & \bc & \mybar{0.5233}  & \kc &     &     &     & \bc & \mybar{0.6993}  \\
				& rs7249750   &     & \pc &     &     & \bc & \mybar{2.3677}  &     & \pc &     &     & \bc & \mybar{2.3502}  &     &     &     &     &     & \mybar{0.6967}  &     & \pc &     &     &     & \mybar{2.1090}  \\
				& 19:45316223 &     & \pc &     &     & \bc & \mybar{2.4536}  &     & \pc &     & \rc & \bc & \mybar{2.4857}  &     &     &     &     &     & \mybar{0.5994}  &     & \pc &     &     &     & \mybar{2.1266}  \\
				& 19:45316330 &     & \pc &     &     & \bc & \mybar{2.3901}  &     & \pc &     & \rc & \bc & \mybar{2.3522}  &     &     &     &     &     & \mybar{0.7109}  &     & \pc &     &     &     & \mybar{2.1061}  \\
				& rs3810141   &     & \pc &     & \rc & \bc & \mybar{2.4601}  &     & \pc &     & \rc & \bc & \mybar{2.3565}  &     &     &     &     &     & \mybar{0.7596}  &     & \pc &     &     &     & \mybar{2.0953}  \\
				& rs3810140   &     & \pc &     & \rc & \bc & \mybar{2.4619}  &     & \pc &     & \rc & \bc & \mybar{2.3571}  &     &     &     &     &     & \mybar{0.7608}  &     & \pc &     &     &     & \mybar{2.0957}  \\
				& rs2968180   &     & \pc &     & \rc & \bc & \mybar{2.8369}  & \kc & \pc & \gc & \rc & \bc & \mybar{2.6676}  & \kc &     &     &     &     & \mybar{1.1770}  &     & \pc &     &     &     & \mybar{1.8398}  \\
				& rs111548706 & \kc & \pc & \gc & \rc & \bc & \mybar{3.1791}  & \kc & \pc & \gc & \rc & \bc & \mybar{2.8133}  &     &     &     &     &     & \mybar{1.1721}  & \kc & \pc &     &     & \bc & \mybar{2.4296}  \\
				& rs1135062   & \kc & \pc &     &     & \bc & \mybar{0.8673}  & \kc &     & \gc &     & \bc & \mybar{0.3313}  &     &     &     &     & \bc & \mybar{0.8898}  & \kc &     & \gc & \rc & \bc & \mybar{1.2600}  \\
				& rs3669      & \kc &     &     &     &     & \mybar{1.2158}  & \kc &     &     &     &     & \mybar{0.6184}  & \kc &     & \gc &     & \bc & \mybar{1.8555}  & \kc &     &     &     & \bc & \mybar{1.9138}  \\
				& 19:45323170 & \kc &     & \gc &     & \bc & \mybar{1.9457}  & \kc &     &     &     & \bc & \mybar{1.3567}  & \kc &     & \gc & \rc & \bc & \mybar{2.7893}  &     &     &     &     & \bc & \mybar{1.1368}  \\
				& rs28399635  &     &     &     &     & \bc & \mybar{0.6250}  &     &     &     &     &     & \mybar{0.0077}  &     &     &     &     &     & \mybar{0.4227}  &     &     &     &     & \bc & \mybar{0.4519}  \\
				& rs28399637  & \kc & \pc & \gc & \rc & \bc & \mybar{5.0766}  & \kc & \pc & \gc & \rc & \bc & \mybar{3.2757}  & \kc &     &     &     & \bc & \mybar{4.3394}  & \kc &     &     &     & \bc & \mybar{2.3330}  \\
				& rs7026      &     &     &     &     & \bc & \mybar{0.3977}  & \kc &     & \gc &     & \bc & \mybar{0.9759}  &     &     &     &     & \bc & \mybar{0.5455}  & \kc &     & \gc &     & \bc & \mybar{0.4153}  \\ \hline
				\multirowcell{40}{PVRL2\\(50/164)}   & rs3810143   &     &     &     &     &     & \mybar{0.0568}  &     &     & \gc &     &     & \mybar{0.8739}  &     &     &     &     &     & \mybar{0.3196}  &     &     &     &     &     & \mybar{0.0028}  \\
				& rs2306149   &     &     &     &     &     & \mybar{0.1418}  &     &     &     &     &     & \mybar{0.5509}  &     &     &     &     &     & \mybar{0.9558}  & \kc &     & \gc &     &     & \mybar{2.3000}  \\
				& rs2972569   &     &     &     &     &     & \mybar{0.0584}  & \kc &     & \gc &     &     & \mybar{1.2822}  &     &     &     &     &     & \mybar{0.5123}  & \kc &     &     &     &     & \mybar{1.6298}  \\
				& rs12974942  &     &     &     &     &     & \mybar{0.1161}  &     &     & \gc &     &     & \mybar{0.7548}  &     &     &     &     &     & \mybar{0.5622}  &     &     &     &     &     & \mybar{0.0219}  \\
				& rs2927469   &     &     &     &     &     & \mybar{0.0391}  &     &     &     &     &     & \mybar{1.0026}  & \kc &     &     &     &     & \mybar{0.0150}  &     &     &     &     &     & \mybar{1.5720}  \\
				& rs2972559   &     &     &     &     &     & \mybar{0.9573}  &     &     & \gc &     &     & \mybar{0.2427}  &     &     &     &     &     & \mybar{2.5134}  &     &     &     &     &     & \mybar{0.1028}  \\
				& rs73050205  &     &     &     &     &     & \mybar{0.8887}  &     &     & \gc &     &     & \mybar{0.2616}  &     &     &     &     &     & \mybar{2.5143}  &     &     &     &     &     & \mybar{0.0707}  \\
				& 19:45356752 &     &     &     &     &     & \mybar{1.4015}  & \kc &     & \gc &     &     & \mybar{0.1564}  &     &     &     &     &     & \mybar{1.9717}  &     &     &     &     &     & \mybar{0.1591}  \\
				& rs35396326  &     &     &     &     &     & \mybar{1.0228}  &     &     & \gc &     &     & \mybar{0.1839}  &     &     &     &     &     & \mybar{2.5206}  &     & \pc &     &     &     & \mybar{0.1790}  \\
				& rs4803763   &     &     &     &     &     & \mybar{0.8644}  &     &     & \gc &     &     & \mybar{0.2685}  &     &     &     &     &     & \mybar{2.5142}  &     & \pc &     &     &     & \mybar{0.0689}  \\
				& rs4803764   &     &     &     &     &     & \mybar{0.8732}  &     &     & \gc &     &     & \mybar{0.2684}  &     &     &     &     &     & \mybar{2.5132}  &     &     &     &     &     & \mybar{0.0650}  \\
				& rs56317818  &     &     &     &     &     & \mybar{0.8706}  &     &     & \gc &     &     & \mybar{0.2718}  &     &     &     &     &     & \mybar{2.6255}  &     &     &     &     &     & \mybar{0.0575}  \\
				& rs12462573  &     &     &     &     &     & \mybar{0.8266}  &     &     & \gc &     &     & \mybar{0.3009}  &     &     &     &     &     & \mybar{2.5528}  &     &     &     &     &     & \mybar{0.0374}  \\
				& rs2972557   & \kc &     &     &     &     & \mybar{0.5298}  &     &     &     &     &     & \mybar{3.5104}  & \kc &     &     &     &     & \mybar{0.5427}  &     &     &     &     &     & \mybar{3.9294}  \\
				& rs12463239  & \kc &     & \gc &     &     & \mybar{1.2425}  &     &     &     &     &     & \mybar{1.6314}  &     &     &     &     &     & \mybar{0.0343}  &     &     &     &     &     & \mybar{0.9865}  \\
				& rs8112526   &     &     &     &     &     & \mybar{1.1821}  &     &     &     &     &     & \mybar{3.9210}  & \kc &     &     &     &     & \mybar{2.0507}  &     &     &     &     &     & \mybar{3.9007}  \\
				& rs3852856   & \kc &     & \gc &     &     & \mybar{2.0895}  & \kc &     & \gc &     &     & \mybar{3.3686}  &     &     &     &     &     & \mybar{0.4751}  & \kc &     & \gc &     &     & \mybar{1.7649}  \\
				& rs3112439   &     &     &     &     &     & \mybar{2.2281}  &     & \pc &     &     &     & \mybar{5.5792}  &     &     &     &     &     & \mybar{3.1366}  &     &     &     &     &     & \mybar{4.6825}  \\
				& rs3112440   & \kc &     & \gc &     &     & \mybar{2.0872}  & \kc & \pc & \gc & \rc &     & \mybar{5.3410}  & \kc & \pc & \gc & \rc &     & \mybar{2.8066}  & \kc &     & \gc &     &     & \mybar{4.8694}  \\
				& rs117877932 & \kc &     & \gc &     &     & \mybar{0.2718}  & \kc &     & \gc &     &     & \mybar{0.2294}  &     &     &     &     &     & \mybar{0.6176}  & \kc &     &     &     &     & \mybar{0.8290}  \\
				& rs11879589  &     &     &     &     &     & \mybar{0.5319}  &     &     &     &     &     & \mybar{0.0128}  &     &     &     &     &     & \mybar{0.4140}  &     & \pc & \gc & \rc &     & \mybar{1.5080}  \\
				& rs3852857   &     &     &     &     &     & \mybar{0.5299}  &     &     &     &     &     & \mybar{0.0128}  &     &     &     &     &     & \mybar{0.4105}  &     & \pc & \gc & \rc &     & \mybar{1.5103}  \\
				& rs395908    &     &     &     &     &     & \mybar{1.3252}  &     &     &     &     &     & \mybar{2.5642}  &     &     &     &     &     & \mybar{1.3123}  &     & \pc &     & \rc &     & \mybar{1.9397}  \\
				& rs4081918   &     &     &     &     &     & \mybar{0.5152}  &     &     &     &     &     & \mybar{0.0067}  &     &     &     &     &     & \mybar{0.3768}  &     & \pc & \gc & \rc &     & \mybar{1.5292}  \\
				& rs79074020  &     &     &     &     &     & \mybar{0.5085}  &     &     &     &     &     & \mybar{0.0178}  &     &     &     &     &     & \mybar{0.3543}  &     & \pc & \gc & \rc &     & \mybar{1.5411}  \\
				& 19:45376044 &     &     &     &     &     & \mybar{1.0236}  &     &     &     &     &     & \mybar{0.2355}  &     &     &     &     &     & \mybar{1.2981}  &     & \pc &     &     &     & \mybar{0.3052}  \\
				& rs519113    &     &     &     &     &     & \mybar{0.4692}  &     &     &     &     &     & \mybar{2.0154}  &     &     &     &     &     & \mybar{0.7368}  &     & \pc &     &     &     & \mybar{3.2749}  \\
				& rs11671274  &     &     &     &     &     & \mybar{0.4626}  &     &     &     &     &     & \mybar{0.0118}  &     &     &     &     &     & \mybar{0.2041}  &     & \pc &     &     &     & \mybar{1.1249}  \\
				& rs11672399  &     &     &     &     &     & \mybar{0.7313}  &     &     &     &     &     & \mybar{0.0025}  &     &     &     &     &     & \mybar{0.2237}  & \kc & \pc & \gc &     &     & \mybar{1.6589}  \\
				& rs369599    &     &     &     &     &     & \mybar{0.9638}  &     & \pc &     &     &     & \mybar{2.3554}  &     &     &     &     &     & \mybar{0.3546}  &     &     &     &     &     & \mybar{0.7190}  \\
				& rs412776    &     &     &     &     &     & \mybar{2.8945}  &     & \pc &     &     &     & \mybar{5.0184}  &     &     &     &     &     & \mybar{2.2935}  &     &     &     &     &     & \mybar{4.3709}  \\
				& 19:45379566 &     &     &     &     &     & \mybar{0.9902}  &     & \pc &     &     &     & \mybar{2.3269}  &     &     &     &     &     & \mybar{0.3512}  &     &     &     &     &     & \mybar{0.6856}  \\
				& rs370705    &     &     &     &     &     & \mybar{0.9759}  &     & \pc &     &     &     & \mybar{2.3138}  &     &     &     &     &     & \mybar{0.3368}  &     &     &     &     &     & \mybar{0.6816}  \\
				& rs385982    &     &     &     &     &     & \mybar{0.9731}  &     & \pc &     &     &     & \mybar{2.3153}  &     &     &     &     &     & \mybar{0.3374}  &     &     &     &     &     & \mybar{0.6842}  \\
				& rs11669338  &     &     &     &     &     & \mybar{2.1823}  &     &     &     &     &     & \mybar{2.1362}  & \kc &     & \gc &     &     & \mybar{4.5319}  &     &     &     &     &     & \mybar{0.6529}  \\
				& rs11673139  &     &     &     &     &     & \mybar{2.1823}  &     &     &     &     &     & \mybar{2.1374}  &     &     & \gc &     &     & \mybar{4.5324}  &     &     &     &     &     & \mybar{0.6545}  \\
				& rs71352237  &     & \pc &     & \rc &     & \mybar{3.0270}  &     & \pc &     & \rc &     & \mybar{5.5205}  &     &     &     &     &     & \mybar{2.4615}  &     & \pc &     & \rc &     & \mybar{4.9586}  \\
				& 19:45383091 &     & \pc &     & \rc &     & \mybar{3.0326}  &     & \pc &     & \rc &     & \mybar{5.5073}  &     &     &     &     &     & \mybar{2.4421}  &     & \pc &     & \rc &     & \mybar{4.9404}  \\
				& rs34224078  &     & \pc &     & \rc &     & \mybar{3.1238}  &     & \pc &     & \rc &     & \mybar{5.6828}  &     &     &     &     &     & \mybar{2.5423}  &     & \pc &     & \rc &     & \mybar{5.1330}  \\
				& rs35879138  &     & \pc &     & \rc &     & \mybar{3.1266}  &     & \pc &     & \rc &     & \mybar{5.6869}  &     &     &     &     &     & \mybar{2.5450}  &     & \pc &     & \rc &     & \mybar{5.1373}  \\
				\multirowcell{10}{} & rs11083749  &     &     &     &     &     & \mybar{0.1248}  &     &     &     &     &     & \mybar{0.7118}  &     &     &     &     &     & \mybar{0.7168}  & \kc & \pc &     &     &     & \mybar{0.6826}  \\
				& rs3745150   & \kc &     & \gc &     &     & \mybar{0.6945}  & \kc &     & \gc &     &     & \mybar{2.0582}  &     &     &     & \rc &     & \mybar{0.7033}  & \kc &     & \gc &     &     & \mybar{0.7161}  \\
				& 19:45386467 &     &     &     &     &     & \mybar{7.2087}  &     &     &     &     &     & \mybar{11.4767} &     & \pc &     & \rc &     & \mybar{12.3805} &     &     &     &     &     & \mybar{11.2765} \\
				& rs12972156  &     &     &     &     &     & \mybar{7.4868}  &     &     &     &     &     & \mybar{13.6269} &     & \pc &     & \rc &     & \mybar{12.7336} &     &     &     &     &     & \mybar{13.2125} \\
				& rs12972970  &     &     &     &     &     & \mybar{7.4876}  &     &     &     &     &     & \mybar{13.6316} &     & \pc &     & \rc &     & \mybar{12.7382} &     &     &     &     &     & \mybar{13.2184} \\
				& rs34342646  &     &     &     &     &     & \mybar{7.4875}  &     &     &     &     &     & \mybar{13.6346} &     & \pc &     & \rc &     & \mybar{12.7429} &     &     &     &     &     & \mybar{13.2230} \\
				& rs283812    &     &     &     &     &     & \mybar{7.2167}  &     &     &     &     &     & \mybar{15.5860} & \kc &     &     &     &     & \mybar{13.8906} &     &     &     &     &     & \mybar{15.0706} \\
				& rs283814    &     &     &     &     &     & \mybar{0.5026}  & \kc &     &     &     &     & \mybar{1.9275}  & \kc &     & \gc &     &     & \mybar{3.0788}  &     &     &     &     &     & \mybar{2.8765} \\
				& rs283815    &     &     &     &     &     & \mybar{6.6893}  &     &     &     & \rc &     & \mybar{13.1772} &     &     &     & \rc &     & \mybar{11.1796} &     &     &     &     &     & \mybar{12.4696} \\
				& rs6857      &     &     &     &     &     & \mybar{7.4645}  &     &     &     & \rc &     & \mybar{16.7899} &     &     &     & \rc &     & \mybar{14.5335} &     &     &     & \rc &     & \mybar{17.6298} \\ \hline
				\multirowcell{15}{TOMM40\\(15/38)}  & rs184017    &     &     &     &     &     & \mybar{6.5911}  &     &     &     & \rc &     & \mybar{13.5901} &     &     &     & \rc &     & \mybar{11.4050} &     &     &     & \rc &     & \mybar{13.1810} \\
				& rs157580    &     &     &     &     &     & \mybar{3.9593}  & \kc &     &     & \rc &     & \mybar{7.0721}  & \kc &     & \gc & \rc & \bc & \mybar{9.1044}  & \kc &     &     & \rc &     & \mybar{10.2890} \\
				& rs2075649   &     &     &     &     &     & \mybar{1.0123}  &     &     &     &     &     & \mybar{2.1800}  &     &     &     & \rc &     & \mybar{0.5978}  &     &     &     &     &     & \mybar{0.7212} \\
				& rs2075650   &     &     &     &     &     & \mybar{7.4179}  &     &     &     &     &     & \mybar{14.1030} &     & \pc &     & \rc & \bc & \mybar{12.8059} &     &     &     &     &     & \mybar{13.6997} \\
				& rs157581    &     &     &     &     &     & \mybar{6.6616}  &     &     &     &     &     & \mybar{13.5959} &     & \pc &     & \rc &     & \mybar{11.4630} &     &     &     &     &     & \mybar{13.2322} \\
				& rs34095326  &     &     &     &     &     & \mybar{5.3435}  &     &     &     &     &     & \mybar{6.1919}  &     & \pc &     & \rc &     & \mybar{9.9105}  &     &     &     &     &     & \mybar{6.2489} \\
				& rs34404554  &     &     &     &     &     & \mybar{7.4144}  &     & \pc &     & \rc &     & \mybar{14.0686} &     & \pc &     & \rc & \bc & \mybar{12.7886} &     &     &     &     &     & \mybar{13.6674} \\
				& rs11556505  &     &     &     &     &     & \mybar{7.4135}  &     & \pc &     & \rc &     & \mybar{14.0612} &     & \pc &     & \rc & \bc & \mybar{12.7845} &     &     &     &     &     & \mybar{13.6604} \\
				& rs157582    &     &     &     &     &     & \mybar{6.5957}  &     & \pc &     & \rc &     & \mybar{13.5386} &     & \pc &     & \rc &     & \mybar{11.3573} &     &     &     &     &     & \mybar{13.1245} \\
				& rs59007384  &     &     &     &     &     & \mybar{7.6850}  &     & \pc &     & \rc &     & \mybar{15.1743} &     & \pc &     & \rc &     & \mybar{12.2690} &     &     &     &     &     & \mybar{14.6344} \\
				& rs11668327  & \kc & \pc & \gc & \rc & \bc & \mybar{6.3772}  & \kc & \pc &     &     &     & \mybar{6.9157}  & \kc & \pc & \gc &     & \bc & \mybar{6.7505}  & \kc &     &     &     &     & \mybar{4.0387}  \\
				& rs118170342 & \kc &     &     &     &     & \mybar{1.7624}  & \kc &     &     &     &     & \mybar{3.1034}  &     &     &     &     &     & \mybar{2.4722}  & \kc &     &     &     &     & \mybar{4.4206}  \\
				& rs35568738  &     &     &     &     &     & \mybar{1.0708}  &     &     &     &     &     & \mybar{0.1818}  & \kc &     & \gc &     &     & \mybar{1.0092}  &     &     &     &     &     & \mybar{0.4404}  \\
				& rs1160984   & \kc &     &     &     &     & \mybar{1.6633}  &     &     &     &     &     & \mybar{0.2869}  & \kc &     & \gc &     &     & \mybar{1.2824}  &     &     &     &     &     & \mybar{0.4915}  \\
				& rs10119     &     &     &     &     &     & \mybar{4.1452}  &     &     &     &     &     & \mybar{9.8079}  &     &     &     &     &     & \mybar{7.8351}  &     &     &     & \rc &     & \mybar{13.0492} \\ \hline
				\multirowcell{5}{APOE\\(5/5)}    & rs440446    &     &     & \gc &     & \bc & \mybar{3.6625}  &     &     & \gc & \rc & \bc & \mybar{5.2200}  &     &     & \gc &     & \bc & \mybar{6.0049}  &     &     & \gc & \rc & \bc & \mybar{6.5638}  \\
				& rs769449    & \kc & \pc & \gc & \rc & \bc & \mybar{13.1079} & \kc & \pc & \gc & \rc & \bc & \mybar{22.4118} & \kc &     & \gc & \rc & \bc & \mybar{16.1028} & \kc & \pc & \gc & \rc & \bc & \mybar{21.0525} \\
				& rs769450    &     &     &     & \rc & \bc & \mybar{2.6047}  &     & \pc & \gc & \rc & \bc & \mybar{5.6348}  &     &     &     & \rc & \bc & \mybar{2.0373}  &     & \pc &     & \rc & \bc & \mybar{3.4491}  \\
				& rs429358    & \kc & \pc & \gc & \rc & \bc & \mybar{13.5662} & \kc & \pc & \gc & \rc & \bc & \mybar{25.2546} & \kc & \pc & \gc & \rc & \bc & \mybar{17.9385} & \kc & \pc & \gc & \rc & \bc & \mybar{25.0369} \\
				& rs7412      & \kc &     & \gc & \rc & \bc & \mybar{1.8479}  & \kc &     & \gc & \rc & \bc & \mybar{3.0122}  & \kc &     & \gc & \rc & \bc & \mybar{3.0474}  & \kc &     & \gc & \rc & \bc & \mybar{4.6691}  \\ \hline
				\multirowcell{14}{APOC1\\(14/14)}   & 19:45417632 &     & \pc &     & \rc & \bc & \mybar{8.3927}  &     &     &     & \rc & \bc & \mybar{13.3937} &     & \pc &     & \rc & \bc & \mybar{10.9990} &     &     &     & \rc & \bc & \mybar{12.5596} \\
				& 19:45417638 &     & \pc &     & \rc & \bc & \mybar{8.3749}  &     &     &     & \rc & \bc & \mybar{13.3586} &     & \pc &     & \rc & \bc & \mybar{10.9658} &     &     &     & \rc & \bc & \mybar{12.5394} \\
				& rs12691088  & \kc &     &     & \rc & \bc & \mybar{3.4966}  & \kc &     & \gc &     & \bc & \mybar{2.4257}  &     & \pc &     & \rc & \bc & \mybar{5.5566}  & \kc &     & \gc &     & \bc & \mybar{1.8214}  \\
				& rs5117      &     &     &     & \rc & \bc & \mybar{8.8625}  &     &     &     & \rc & \bc & \mybar{15.8799} &     & \pc &     & \rc & \bc & \mybar{11.5395} &     &     &     & \rc & \bc & \mybar{14.7969} \\
				& rs3826688   &     &     &     & \rc &     & \mybar{2.1575}  &     &     &     &     &     & \mybar{3.3487}  &     &     &     & \rc &     & \mybar{3.6469}  &     &     &     & \rc &     & \mybar{2.9339}  \\
				& rs73052335  & \kc &     & \gc & \rc & \bc & \mybar{12.3385} &     &     &     & \rc & \bc & \mybar{20.4728} &     &     &     & \rc & \bc & \mybar{15.5834} &     &     & \gc & \rc & \bc & \mybar{19.9258} \\
				& rs3925681   & \kc &     & \gc & \rc & \bc & \mybar{4.8058}  &     &     &     &     & \bc & \mybar{7.4320}  &     &     &     &     &     & \mybar{3.8688}  & \kc &     & \gc & \rc & \bc & \mybar{9.8087}  \\
				& rs150966173 &     &     &     &     &     & \mybar{0.9923}  &     &     &     &     &     & \mybar{2.5309}  &     &     &     &     &     & \mybar{1.6387}  &     &     &     & \rc & \bc & \mybar{3.3859}  \\
				& rs12721046  &     &     &     & \rc & \bc & \mybar{11.2256} &     &     &     & \rc & \bc & \mybar{19.1327} & \kc &     &     & \rc & \bc & \mybar{14.8090} &     &     &     & \rc & \bc & \mybar{18.4076} \\
				& rs12721056  &     &     &     &     & \bc & \mybar{4.1169}  &     &     &     &     & \bc & \mybar{7.2385}  &     &     &     &     &     & \mybar{3.3861}  & \kc &     & \gc & \rc & \bc & \mybar{7.9466}  \\
				& rs484195    &     &     &     &     &     & \mybar{2.8087}  &     &     &     &     & \bc & \mybar{4.5296}  &     &     &     &     & \bc & \mybar{4.4419}  &     &     &     & \rc & \bc & \mybar{4.0451}  \\
				& 19:45421972 &     &     &     &     &     & \mybar{3.8770}  & \kc &     &     & \rc & \bc & \mybar{7.2418}  & \kc &     &     &     &     & \mybar{3.9087}  & \kc &     & \gc & \rc & \bc & \mybar{7.8283}  \\
				& rs12721051  &     &     &     & \rc & \bc & \mybar{11.6336} &     &     &     & \rc & \bc & \mybar{21.6248} &     &     &     &     & \bc & \mybar{15.6163} &     &     &     & \rc & \bc & \mybar{22.2684} \\
				& rs1064725   &     &     &     &     & \bc & \mybar{2.2981}  &     &     &     &     &     & \mybar{2.0402}  &     &     &     &     &     & \mybar{1.4059}  & \kc &     &     &     &     & \mybar{0.3741}  \\ \hline
				\multirowcell{7}{APOC4\\(7/7)}   & 19:45445860 & \kc & \pc &     & \rc & \bc & \mybar{0.2335}  &     &     & \gc & \rc & \bc & \mybar{1.3186}  & \kc &     &     &     & \bc & \mybar{0.3865}  &     &     &     & \rc & \bc & \mybar{0.6790}  \\
				& 19:45446261 & \kc & \pc & \gc & \rc & \bc & \mybar{3.8279}  & \kc & \pc & \gc & \rc & \bc & \mybar{9.8048}  & \kc &     &     & \rc & \bc & \mybar{2.5898}  & \kc & \pc & \gc & \rc & \bc & \mybar{7.7119}  \\
				& 19:45446271 &     &     &     &     & \bc & \mybar{0.8625}  &     &     &     &     & \bc & \mybar{0.8353}  & \kc &     &     & \rc & \bc & \mybar{0.9413}  &     &     &     &     & \bc & \mybar{1.5025}  \\
				& rs5157      &     &     &     &     & \bc & \mybar{1.1912}  &     &     &     &     & \bc & \mybar{0.3194}  &     &     &     &     & \bc & \mybar{0.5001}  &     &     &     &     & \bc & \mybar{1.2607}  \\
				& rs5158      &     &     &     &     & \bc & \mybar{0.0375}  & \kc &     & \gc &     & \bc & \mybar{1.5699}  &     &     &     &     & \bc & \mybar{0.2380}  & \kc &     &     &     & \bc & \mybar{0.9237}  \\
				& rs1132899   &     &     &     &     & \bc & \mybar{1.2890}  &     &     &     &     & \bc & \mybar{0.4523}  &     &     &     &     & \bc & \mybar{0.7862}  &     &     &     &     & \bc & \mybar{1.4671}  \\
				& rs5167      &     & \pc &     & \rc & \bc & \mybar{1.4347}  &     & \pc &     &     & \bc & \mybar{1.8962}  &     &     &     &     & \bc & \mybar{0.6398}  & \kc & \pc &     & \rc & \bc & \mybar{2.8189}  \\ \hline
				\multirowcell{26}{EXOC3L2\\(26/88)} & 19:45715976 & \kc & \pc & \gc & \rc &     & \mybar{3.4808}  & \kc & \pc & \gc & \rc &     & \mybar{1.3313}  & \kc & \pc & \gc & \rc &     & \mybar{1.0972}  & \kc & \pc & \gc & \rc &     & \mybar{2.0754}  \\
				& 19:45716192 &     & \pc & \gc & \rc &     & \mybar{1.8433}  & \kc & \pc & \gc & \rc &     & \mybar{1.4002}  &     & \pc &     & \rc &     & \mybar{0.1588}  & \kc & \pc & \gc & \rc &     & \mybar{1.8926}  \\
				& 19:45716197 &     & \pc & \gc & \rc &     & \mybar{1.7716}  &     & \pc &     & \rc &     & \mybar{0.7414}  &     &     &     &     &     & \mybar{0.0801}  &     & \pc &     & \rc &     & \mybar{1.4170}  \\
				& 19:45716678 &     & \pc &     &     &     & \mybar{0.1425}  &     & \pc &     &     &     & \mybar{0.3373}  & \kc &     &     &     &     & \mybar{1.2363}  &     & \pc &     &     &     & \mybar{0.6786}  \\
				& 19:45717615 &     & \pc & \gc & \rc &     & \mybar{1.8636}  &     & \pc & \gc & \rc &     & \mybar{2.4635}  &     &     &     &     &     & \mybar{0.6285}  &     & \pc &     & \rc &     & \mybar{1.6887}  \\
				& 19:45718624 &     & \pc & \gc & \rc &     & \mybar{1.8734}  &     & \pc & \gc & \rc &     & \mybar{2.4718}  &     &     &     &     &     & \mybar{0.6184}  &     & \pc &     & \rc &     & \mybar{1.6884}  \\
				& 19:45719065 & \kc &     &     &     &     & \mybar{0.9007}  &     &     &     &     &     & \mybar{0.4738}  &     &     &     &     &     & \mybar{0.3541}  &     &     &     &     &     & \mybar{0.1623}  \\
				& 19:45721596 & \kc &     & \gc & \rc &     & \mybar{2.0318}  & \kc &     & \gc &     &     & \mybar{2.5855}  &     &     &     &     &     & \mybar{0.6111}  & \kc &     &     &     &     & \mybar{1.8777}  \\
				& 19:45723379 & \kc &     &     &     &     & \mybar{1.1438}  &     &     &     &     &     & \mybar{0.0876}  &     &     &     &     &     & \mybar{0.2857}  &     &     &     &     &     & \mybar{0.8241}  \\
				& 19:45724044 &     &     &     &     &     & \mybar{1.1222}  & \kc &     &     &     &     & \mybar{0.0220}  & \kc &     &     &     &     & \mybar{0.0262}  &     &     &     &     &     & \mybar{0.2408}  \\
				& rs10405194  &     &     &     &     &     & \mybar{0.0231}  &     &     &     &     &     & \mybar{0.2934}  &     & \pc &     &     &     & \mybar{1.6704}  &     &     &     &     &     & \mybar{0.1134}  \\
				& rs10403626  &     &     &     &     &     & \mybar{0.0230}  &     &     &     &     &     & \mybar{0.2930}  &     & \pc &     &     &     & \mybar{1.6707}  &     &     &     &     &     & \mybar{0.1136}  \\
				& rs10409909  &     &     &     &     &     & \mybar{0.0228}  &     &     &     &     &     & \mybar{0.2921}  &     & \pc &     &     &     & \mybar{1.6717}  &     &     &     &     &     & \mybar{0.1143}  \\
				& rs10411314  &     &     &     &     &     & \mybar{0.0227}  &     &     &     &     &     & \mybar{0.2917}  &     & \pc &     &     &     & \mybar{1.6720}  &     &     &     &     &     & \mybar{0.1146}  \\
				& rs10410003  &     &     &     &     &     & \mybar{0.0224}  &     &     &     &     &     & \mybar{0.2912}  &     & \pc &     &     &     & \mybar{1.6706}  &     &     &     &     &     & \mybar{0.1152}  \\
				& rs10410561  &     &     &     &     &     & \mybar{0.0208}  &     &     &     &     &     & \mybar{0.2894}  &     & \pc &     &     &     & \mybar{1.6709}  &     &     &     &     &     & \mybar{0.1160}  \\
				& rs10411743  &     &     &     &     &     & \mybar{0.0115}  &     &     &     &     &     & \mybar{0.2775}  &     & \pc &     & \rc &     & \mybar{1.6718}  &     &     &     &     &     & \mybar{0.1214}  \\
				& 19:45728355 &     &     &     &     &     & \mybar{0.0108}  &     &     &     &     &     & \mybar{0.2767}  &     &     &     & \rc &     & \mybar{1.6720}  &     &     &     &     &     & \mybar{0.1218}  \\
				& rs10412154  &     &     &     &     &     & \mybar{0.0100}  &     &     &     &     &     & \mybar{0.2757}  &     & \pc &     & \rc &     & \mybar{1.6721}  &     &     &     &     &     & \mybar{0.1223}  \\
				& 19:45728440 &     &     &     &     &     & \mybar{0.0247}  &     &     &     &     &     & \mybar{0.2412}  &     &     &     & \rc &     & \mybar{1.6836}  &     &     &     &     &     & \mybar{0.1767}  \\
				& rs346761    &     &     &     &     &     & \mybar{0.0045}  &     &     &     &     &     & \mybar{0.2752}  &     &     &     & \rc &     & \mybar{1.6777}  &     &     &     &     &     & \mybar{0.1237}  \\
				& rs10412614  &     &     &     &     &     & \mybar{0.0064}  &     &     &     &     &     & \mybar{0.2713}  &     &     &     & \rc &     & \mybar{1.6725}  &     &     &     &     &     & \mybar{0.1242}  \\
				& rs346772    &     &     &     &     &     & \mybar{0.6101}  &     &     &     &     &     & \mybar{0.0843}  & \kc &     & \gc &     &     & \mybar{2.8401}  &     &     &     &     &     & \mybar{0.2026}  \\
				& 19:45733782 & \kc &     &     &     &     & \mybar{1.3478}  & \kc &     &     &     &     & \mybar{0.2223}  & \kc &     &     &     &     & \mybar{0.0059}  & \kc &     &     &     &     & \mybar{0.1435}  \\
				& rs8109472   &     & \pc &     & \rc &     & \mybar{1.4264}  &     &     &     &     &     & \mybar{0.1452}  &     & \pc &     &     &     & \mybar{1.1101}  &     &     &     &     &     & \mybar{0.1373}  \\
				& rs346750    &     & \pc &     & \rc &     & \mybar{1.4303}  &     &     &     &     &     & \mybar{0.1707}  &     & \pc &     &     &     & \mybar{1.0525}  &     &     &     &     &     & \mybar{0.1772}  \\ \hline
				\multirowcell{16}{CD33\\(16/16)}    & rs12459419  & \kc &     & \gc & \rc & \bc & \mybar{1.0907}  &     &     &     &     & \bc & \mybar{0.4046}  &     &     &     &     & \bc & \mybar{1.7706}  &     &     &     &     & \bc & \mybar{0.9117}  \\
				& rs2455069   & \kc &     &     &     & \bc & \mybar{0.8196}  & \kc &     &     &     & \bc & \mybar{0.0588}  & \kc &     & \gc &     & \bc & \mybar{0.2312}  & \kc &     &     &     & \bc & \mybar{0.1462}  \\
				& rs7245846   &     &     &     &     &     & \mybar{0.4353}  &     & \pc &     &     & \bc & \mybar{1.0054}  &     & \pc &     &     & \bc & \mybar{2.1357}  &     & \pc &     &     & \bc & \mybar{1.2773}  \\
				& 19:51734857 &     &     &     &     &     & \mybar{0.3602}  &     & \pc &     &     & \bc & \mybar{1.1170}  &     & \pc &     &     & \bc & \mybar{2.2893}  &     & \pc &     &     & \bc & \mybar{1.4001}  \\
				& 19:51735023 &     & \pc &     &     & \bc & \mybar{0.7299}  &     & \pc &     &     &     & \mybar{0.6667}  &     & \pc &     &     & \bc & \mybar{2.2540}  &     & \pc &     &     & \bc & \mybar{1.2577}  \\
				& rs33978622  &     & \pc &     &     & \bc & \mybar{0.7307}  &     & \pc &     &     &     & \mybar{0.6965}  &     & \pc &     &     & \bc & \mybar{2.2992}  &     & \pc &     &     & \bc & \mybar{1.3059}  \\
				& rs34813869  &     & \pc &     &     &     & \mybar{0.3891}  &     & \pc &     &     & \bc & \mybar{1.1139}  &     & \pc &     &     & \bc & \mybar{2.4610}  &     & \pc &     &     & \bc & \mybar{1.5210}  \\
				& rs1354106   &     & \pc &     &     & \bc & \mybar{0.3100}  &     & \pc &     &     & \bc & \mybar{1.2263}  &     & \pc & \gc &     & \bc & \mybar{2.4395}  &     & \pc &     &     & \bc & \mybar{1.5045}  \\
				& rs35112940  & \kc & \pc &     & \rc & \bc & \mybar{1.3723}  &     &     &     &     & \bc & \mybar{0.7697}  & \kc &     & \gc & \rc & \bc & \mybar{1.9413}  & \kc &     &     & \rc & \bc & \mybar{1.2327}  \\
				& rs10409348  & \kc &     & \gc & \rc & \bc & \mybar{0.3210}  & \kc &     & \gc & \rc & \bc & \mybar{2.0647}  & \kc & \pc & \gc & \rc & \bc & \mybar{4.0526}  & \kc & \pc & \gc & \rc & \bc & \mybar{2.8254}  \\
				& rs146995981 & \kc &     & \gc &     & \bc & \mybar{0.2094}  & \kc &     &     &     & \bc & \mybar{0.0575}  & \kc &     & \gc & \rc & \bc & \mybar{0.3913}  & \kc &     &     &     & \bc & \mybar{0.1173}  \\
				& rs1399839   &     &     &     &     & \bc & \mybar{0.0542}  &     &     &     &     & \bc & \mybar{0.9633}  &     & \pc &     & \rc & \bc & \mybar{1.5430}  &     &     &     &     & \bc & \mybar{0.6127}  \\
				& rs273653    &     &     &     &     & \bc & \mybar{0.0476}  &     &     &     &     & \bc & \mybar{0.9540}  &     & \pc &     & \rc & \bc & \mybar{1.5763}  &     &     &     &     & \bc & \mybar{0.5891}  \\
				& rs273652    &     &     &     &     & \bc & \mybar{0.1969}  &     &     &     &     & \bc & \mybar{0.5591}  & \kc & \pc & \gc & \rc & \bc & \mybar{3.6421}  & \kc &     &     &     & \bc & \mybar{0.6791}  \\
				& rs75773078  & \kc & \pc & \gc & \rc & \bc & \mybar{1.9834}  & \kc &     & \gc &     & \bc & \mybar{1.2255}  & \kc &     & \gc & \rc & \bc & \mybar{0.0023}  & \kc &     &     &     & \bc & \mybar{0.1933}  \\
				& rs1803254   & \kc & \pc &     & \rc & \bc & \mybar{0.5649}  & \kc &     &     & \rc & \bc & \mybar{0.6272}  & \kc & \pc & \gc & \rc & \bc & \mybar{2.8029}  & \kc &     &     &     & \bc & \mybar{0.0603} 
			\end{longtable}
		}
	\end{landscape}
	\clearpage
}

\end{document}